\documentclass[journal]{IEEEtran}
\usepackage{amsmath,amsfonts}
\usepackage{algorithmic}
\usepackage{algorithm}
\usepackage{array}
\usepackage[caption=false,font=normalsize,labelfont=sf,textfont=sf, labelformat=simple, labelsep=space]{subfig}
\usepackage{textcomp}
\usepackage{stfloats}
\usepackage{url}
\usepackage{verbatim}
\usepackage{graphicx}
\usepackage{cite}
\usepackage{newfloat}
\usepackage{listings}
\usepackage{amsmath}
\usepackage{booktabs}
\usepackage{multirow}
\usepackage{tabularx}
\usepackage{amssymb}
\usepackage{xcolor}
\usepackage{colortbl}

\begin{document}

\title{TraceMark-LDM: Authenticatable Watermarking for Latent Diffusion Models via Binary-Guided Rearrangement}

% \author{IEEE Publication Technology,~\IEEEmembership{Staff,~IEEE,}
\author {
    % Authors
    Wenhao Luo,
    Zhangyi Shen,
    Ye Yao,
    Feng Ding,
    Guopu Zhu,
    Weizhi Meng,~\IEEEmembership{Senior Member,~IEEE}
% \affiliations {
%     % Affiliations
%     \textsuperscript{\rm 1}Hangzhou Dianzi University\\
%     \textsuperscript{\rm 2}Nanchang University\\
%     \textsuperscript{\rm 3}Harbin Institute of Technology\\
%     \textsuperscript{\rm 4}Technical University of Denmark\\
%     luowenhao@hdu.edu.cn, shenzhangyi@hdu.edu.cn, yaoye@hdu.edu.cn, fengding@ncu.edu.cn, guopu.zhu@hit.edu.cn, weme@dtu.dk
% }
        % <-this % stops a space
\thanks{
\emph{(Corresponding author: Zhangyi Shen)}

W.~Luo, Z.~Shen, Y.~Yao are with the Hangzhou Dianzi University, Hangzhou, ZJ 310018, China (E-mail:  luowenhao@hdu.edu.cn, shenzhangyi@hdu.edu.cn, yaoye@hdu.edu.cn).

F.~Ding is with Nanchang University, Nanchang, JX 330031, China (E-mail: fengding@ncu.edu.cn).

G.~Zhu is with the School of Computer Science and Technology, Harbin Institute of Technology, Harbin 150001, China (E-mail: guopu.zhu@hit.edu.cn).

W.~Meng is with the School of Computing and Communications, Lancaster University, United Kingdom (E-mail: weizhi.meng@ieee.org).}% <-this % stops a space
}

% The paper headers
% \markboth{Journal of \LaTeX\ Class Files,~Vol.~14, No.~8, August~2021}%
% {Shell \MakeLowercase{\textit{et al.}}: A Sample Article Using IEEEtran.cls for IEEE Journals}

% \IEEEpubid{0000--0000/00\$00.00~\copyright~2021 IEEE}
% Remember, if you use this you must call \IEEEpubidadjcol in the second
% column for its text to clear the IEEEpubid mark.

\maketitle

\begin{abstract}
Image generation algorithms are increasingly integral to diverse aspects of human society, driven by their practical applications. However, insufficient oversight in artificial Intelligence generated content (AIGC) can facilitate the spread of malicious content and increase the risk of copyright infringement. Among the diverse range of image generation models, the Latent Diffusion Model (LDM) is currently the most widely used, dominating the majority of the Text-to-Image model market. Currently, most attribution methods for LDMs rely on directly embedding watermarks into the generated images or their intermediate noise, a practice that compromises both the quality and the robustness of the generated content. To address these limitations, we introduce TraceMark-LDM, an novel algorithm that integrates watermarking to attribute generated images while guaranteeing non-destructive performance. Unlike current methods, TraceMark-LDM leverages watermarks as guidance to rearrange random variables sampled from a Gaussian distribution. To mitigate potential deviations caused by inversion errors, the small absolute elements are grouped and rearranged. Additionally, we fine-tune the LDM encoder to enhance the robustness of the watermark. Experimental results show that images synthesized using TraceMark-LDM exhibit superior quality and attribution accuracy compared to state-of-the-art (SOTA) techniques. Notably, TraceMark-LDM demonstrates exceptional robustness against various common attack methods, consistently outperforming SOTA methods.
\end{abstract}


\section{Introduction}
\IEEEPARstart{T}{ext}-to-Image generation enables creating images based on descriptive text prompts \cite{xu2018attngan, qiao2019mirrorgan, song2019generative, song2020score}. This technology has significantly impacted various fields, such as art, design, and marketing \cite{frolov2021adversarial}. Currently, the Latent Diffusion Model (LDM) \cite{rombach2022high} is particularly notable among generative models due to its efficiency in generating high-quality images \cite{fan2024synthesizing}. However, the widespread accessibility of this technology \cite{dhariwal2021diffusion, nichol2021improved, gu2022vector, ho2022classifier} allows the general public to generate large volumes of synthetic data, which may compromise the reliability of images as indispensable carriers of information \cite{guo2023aigc}. To mitigate this concern, it is essential to implement methods for the attribution of synthesized images.
 
Watermarking has recently emerged as a promising approach for achieving this objective \cite{fernandez2023stable, NEURIPS2023_b54d1757, liu2042aigcreview, yuan2024watermarking, yang2024gaussian,li2024screen,he2020high,qin2023print,zhong2020automated,fang2022end}. In the literature, numerous watermarking-based algorithms have been developed to authenticate images synthesized by LDMs. These algorithms incorporate the watermark into the generated image in various ways, with a specific watermark decoder used to verify the embedded watermark as well as determine the image's attribution. Broadly, these methods can be categorized into three types: post-generation watermarking, in-generation watermarking, and initial noise sampling watermarking.
\begin{figure}[t]
    \centering
    \includegraphics[width=\linewidth]{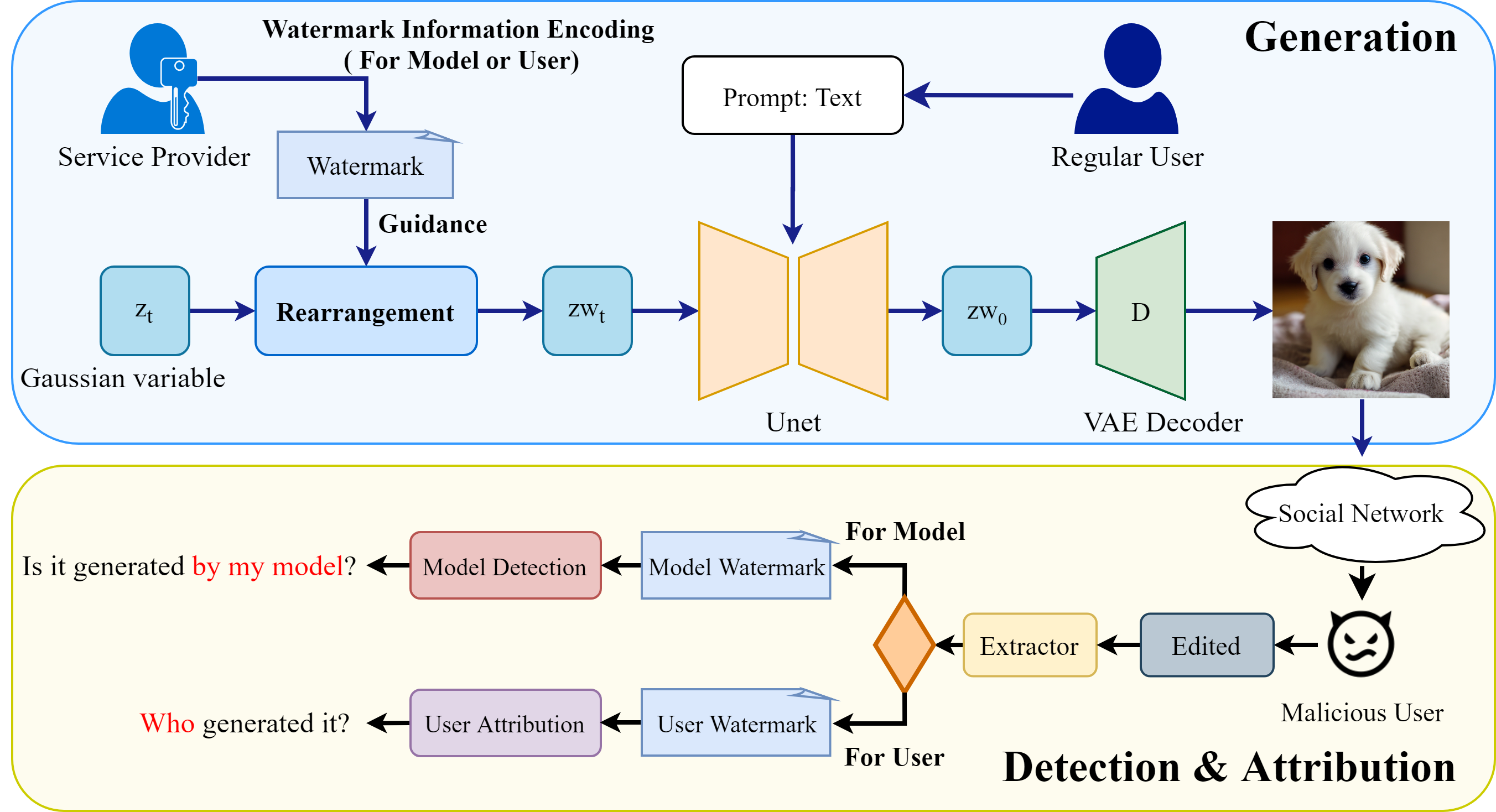}
    \caption{Application scenarios for TraceMark-LDM. The service provider encodes the regular user's information into a watermark and hides the watermark in the generated image. For different cases, watermarks can be designed either to identify whether a given image was generated by the service model (model detection) or to trace the user who generated a specific image (user attribution). When the image is tampered with or infringed upon by malicious users, the extracted watermark can facilitate model detection and user attribution, thereby safeguard the copyright of the regular user.}
    \label{fig:fig1}
\end{figure}
Post-generation watermarking \cite{cox2007digital, Zhang2019RobustIV, yin2022neural, bui2022repmix} involves embedding watermark information into synthesized images after their generation. However, in practical application, the embedded watermarks often incorporate the trademark information of the model owner and user identity details. When the volume of embedded watermark information is substantial, it might reduce the quality of the image. In-generation watermarking refers to the embedding of watermarks during the processes of denoising in the latent space or decoding into the pixel space. For example, AquaLoRA \cite{feng2024aqualora} embeds watermarks during the denoising process by fine-tuning the U-Net structure. In addition, some researchers \cite{fernandez2023stable,cui2023diffusionshield,xiong2023flexible,meng2024latent} have processed denoising latent variables by watermark embedding module or fine-tuning a pre-trained variant autoencoder (VAE) \cite{Kingma2013AutoEncodingVB} decoder to embed watermark information. However, this approach requires more computational resources for training.

Both of the aforementioned algorithms may compromise the quality of generated images, as they directly or semi-directly disturb the proper distribution of the desired output. Furthermore, they are vulnerable to diffusion-based attacks \cite{zhao2023invisible} which further limit their robustness. In contrast, initial noise sampling watermarking methods mitigate this issue by embedding the watermark into the initial latent noise map, avoiding direct interference with the generation outcomes. Their compatibility with any diffusion model, without the need for retraining, makes them particularly versatile and efficient. However, despite these advantages, these methods have some limitations. Tree Ring \cite{NEURIPS2023_b54d1757} embeds a predefined key into the Fourier transform of the initial latent noise, causing the noise to deviate from a Gaussian distribution, thereby decreasing the overall performance of the model. Gaussian Shading \cite{yang2024gaussian} converts watermark information into the initial latent noise through distribution-preserving sampling. However, the encryption involved in this process introduces substantial computational overhead, thereby slowing down the generation speed of the original generative model \cite{lin2024detecting}.

To address various issues in existing LDM watermarking methods, we propose TraceMark-LDM, a non-destructive watermarking scheme designed specifically for LDMs. Fig. \ref{fig:fig1} illustrates the application scenarios of TraceMark-LDM. This method leverages watermark information to guide the rearrangement of a random variable sampled from a Gaussian distribution. The rearrangement is performed according to the direction of its element values to construct initial latent noise for generating watermarked images. During the extraction phase, the watermarked image is mapped to the latent space via a VAE encoder, and the initial watermarked latent noise is recovered through DDIM inversion. Additionally, we enhance the robustness of watermarked images against various corruptions and perturbations by fine-tuning the LDM encoder. In summary, the contributions of this paper are as follows:

\begin{itemize}
    \item This study proposes a general watermarking approach for LDMs that embeds watermarks through the rearrangement of Gaussian distribution samples. This method preserves the LDM's high performance in generating while reducing watermark extraction errors.
    \item We also developed a loss function to ensure that the inverted latent variables are as close as possible to the original ones. This fine-tuning of the LDM encoder reduces perturbation errors. This enhancement significantly enhances the robustness of watermark extraction under various attack scenarios.
    \item We conducted extensive experiments under various attack scenarios with different levels of intensity to comprehensively validate the effectiveness and robustness of the TraceMark-LDM. Additionally, ablation studies were performed to further confirm the efficacy of fine-tuning the LDM encoder.
\end{itemize}

The remainder of this paper is organized as follows. Section II reviews related work. Section III presents the problem statement and background. Section IV discusses the motivation behind our method and outlines the proposed approach. Section V details the experimental setup and results. Finally, the conclusion is given in Section VI.

\begin{table*}[!ht]
\centering
\caption{Qualitative comparison of different watermarking methods.}
\label{tab:qualitative_comparison}
\begin{tabular}{@{}lccccc@{}}
\toprule
\textbf{Method} & \textbf{Performance Lossless} & \textbf{Multi-bit Support} & \textbf{High Robustness} & \textbf{Flexibility} & \textbf{Low Computational Overhead} \\ \midrule
Post-generation \cite{Zhang2019RobustIV} & \(\times\) & \(\checkmark\) & \(\times\) & \(\checkmark\) & \(\checkmark\) \\
Stable Signature \cite{fernandez2023stable} & \(\times\) & \(\checkmark\) & \(\times\) & \(\times\) & \(\checkmark\) \\
Tree-Ring \cite{NEURIPS2023_b54d1757} & \(\times\) & \(\times\) & \(\checkmark\) & \(\checkmark\) & \(\checkmark\) \\
Gaussian Shading \cite{yang2024gaussian} & \(\checkmark\) & \(\checkmark\) & \(\checkmark\) & \(\checkmark\) & \(\times\)  \\
Latent Watermark \cite{meng2024latent} & \(\times\) & \(\checkmark\) & \(\checkmark\) & \(\checkmark\) & \(\checkmark\) \\
\textbf{Ours} & \(\checkmark\) & \(\checkmark\) & \(\checkmark\) & \(\checkmark\) & \(\checkmark\) \\ 
\bottomrule
\end{tabular}
\end{table*}

\section{Related Work}

\subsection{Diffusion Models}
The prototype diffusion model was introduced by Sohl-Dickstein et al. \cite{sohl2015deep}. They developed a deep unsupervised learning algorithm grounded in principles from non-equilibrium statistical physics. This algorithm learns to generate samples matching the target distribution through a forward and a backward Markov chain; however, it suffers from high computational complexity and training costs. Additionally, the quality of generated samples is often suboptimal. Ho et al. \cite{ho2020denoising} then proposed the Denoising Diffusion Probabilistic Model (DDPM), which enhances stability and diversity in sample generation by employing variational inference and a constrained noise schedule strategy. Despite these improvements, the generation speed remains slow. To address this limitation, Song et al. \cite{song2020denoising} developed the Deterministic Denoising Diffusion Implicit Model (DDIM), which redefines the single-step backward diffusion process and enables skipping steps during denoising. This approach facilitates the generation of high-quality images with fewer steps. Furthermore, Rombach et al. \cite{rombach2022high} proposed the Latent Diffusion Model (LDM), which transitions the diffusion process to a lower-dimensional latent space. By incorporating a pre-trained auto-encoder, LDM reduces computational complexity, while also supporting high-resolution image generation. This algorithm significantly improves generation efficiency and sample quality.

\subsection{Deep Learning Watermarking}
As an pioneering deep learning watermarking framework, HiDDeN \cite{zhu2018hidden} achieves imperceptible watermark embedding in images through joint training of encoder and decoder networks, demonstrating remarkable robustness against distortions including blurring, cropping, and compression. RivaGAN \cite{Zhang2019RobustIV} employs a GAN-based architecture enhanced with attention mechanisms and adversarial networks to optimize robustness against distortion and compression artifacts. To improve resistance against JPEG compression, MBRS \cite{jia2021mbrs} introduces a novel end-to-end training framework that incorporates randomized real JPEG compression, simulated JPEG distortion, and noise-free layers as adversarial training components. This approach is further enhanced by integrating Squeeze-and-Excitation (SE) modules for optimized feature extraction and message processors for efficient information embedding. Concurrently, CIN \cite{cin2022} combines reversible and non-reversible mechanisms, where reversible modules ensure high concealment capability while non-reversible components significantly enhance robustness against intensive noise attacks.

\subsection{Watermarking in Generative Models}

Detecting and tracing content generated by generative models can be accomplished by covertly embedding digital watermarking information into the generated content. For example, Stable Diffusion employs traditional digital image watermarking techniques\cite{van1994digital, badran2009multiple,mohammed2014robust,qin2017fragile,fan2021multiple, kang2003dwt,cox2007digital, Zhang2019RobustIV, fang2019robust,wang2022dtcwt,huan2021exploring}, including DwtDct and DwtDctSvd, which can be directly applied to generated images. Additionally, deep learning-based methods can also be used to watermark generated images. In recent years, researchers \cite{xiong2023flexible} have focused on developing specialized watermarking techniques for diffusion models, aiming to improve watermark extraction rates and robustness. Fernandez et al. \cite{fernandez2023stable} proposed a method that fine-tunes the LDM decoder and uses a pre-trained watermark extractor to extract watermarks from watermarked images. However, these methods modify the host image during watermark embedding. When the embedding perturbations are weak, their resistance to distortions remains low; whereas stronger perturbations enhance robustness at the cost of degraded image quality.

To address the common conflict between image quality and watermark robustness in existing watermarking methods, Meng et al. \cite{meng2024latent} designed Latent Watermark, a unified framework for watermark embedding and detection within the latent space, and proposed a progressive training strategy. This approach weakens the direct conflict between robustness and generation quality, thereby alleviating their contradiction. However, the image quality still faces challenges when more bits of watermarks are embedded. Wen et al. \cite{NEURIPS2023_b54d1757} introduced an approach to enhance resistance to corruptions and perturbations by embedding a specially designed pattern, referred to as the key into the Fourier transform domain of the initial latent noise. However, it only supports single-bit watermarking and cannot authenticate the generated content. Additionally, it disrupts the Gaussian distribution of the initial noise, which limits the randomness of the sampling process and sacrifices the performance of the original generative model. In an effort to resolve these issues, Yang et al. \cite{yang2024gaussian} converts watermark information into the initial noise for the generative model through steps such as watermark diffusion, watermark randomization, and distribution-preserving sampling. This approach embeds watermarks without impairing image quality. Nevertheless, the use of Chacha20 encryption during the watermark randomization process is time-consuming, potentially resulting in a decrease in the speed of generation. 

To address the limitations identified in the aforementioned studies, we propose TraceMark-LDM. Table \ref{tab:qualitative_comparison} presents a qualitative comparison between our methods with others. Overall, our approach achieves comprehensive optimality.

\section{Problem Statement and Background}
\subsection{Scenario Description}
As shown in Fig. \ref{fig:fig1}, the image generation and watermarking mechanisms involve the service provider, regular users, and malicious users. The service provider serves as the central entity, responsible for developing and maintaining the image generation model. It offers the API for target users and embeds watermarks to safeguard the copyright of generated content. Regular users are legitimate individuals who adhere to the service agreement and utilize the image generation model by providing text descriptions or other forms of input. They can generate images using the API provided by the service provider. Malicious users are those who attempt to bypass watermark protection mechanisms to generate illicit content or tamper with others’ generated content to commit infringement. These users employ various methods, such as noise addition, noise removal, or image compression, to remove or disguise the watermark and evade detection. Their primary objective is to generate unauthorized images or use generated images for infringing purposes. 

To effectively safeguard the copyright of generated images and attribute illicit content to specific users, the service provider must implement watermarking mechanisms, which can be divided into two categories based on functionality: \textbf{detection watermarking} and \textbf{attribution watermarking}.

\begin{figure*}[ht]

\centering
    \centering
    \subfloat[\label{fig:denoising_inversion_process}]{% <-- 子图标签
        \includegraphics[width=0.35\textwidth]{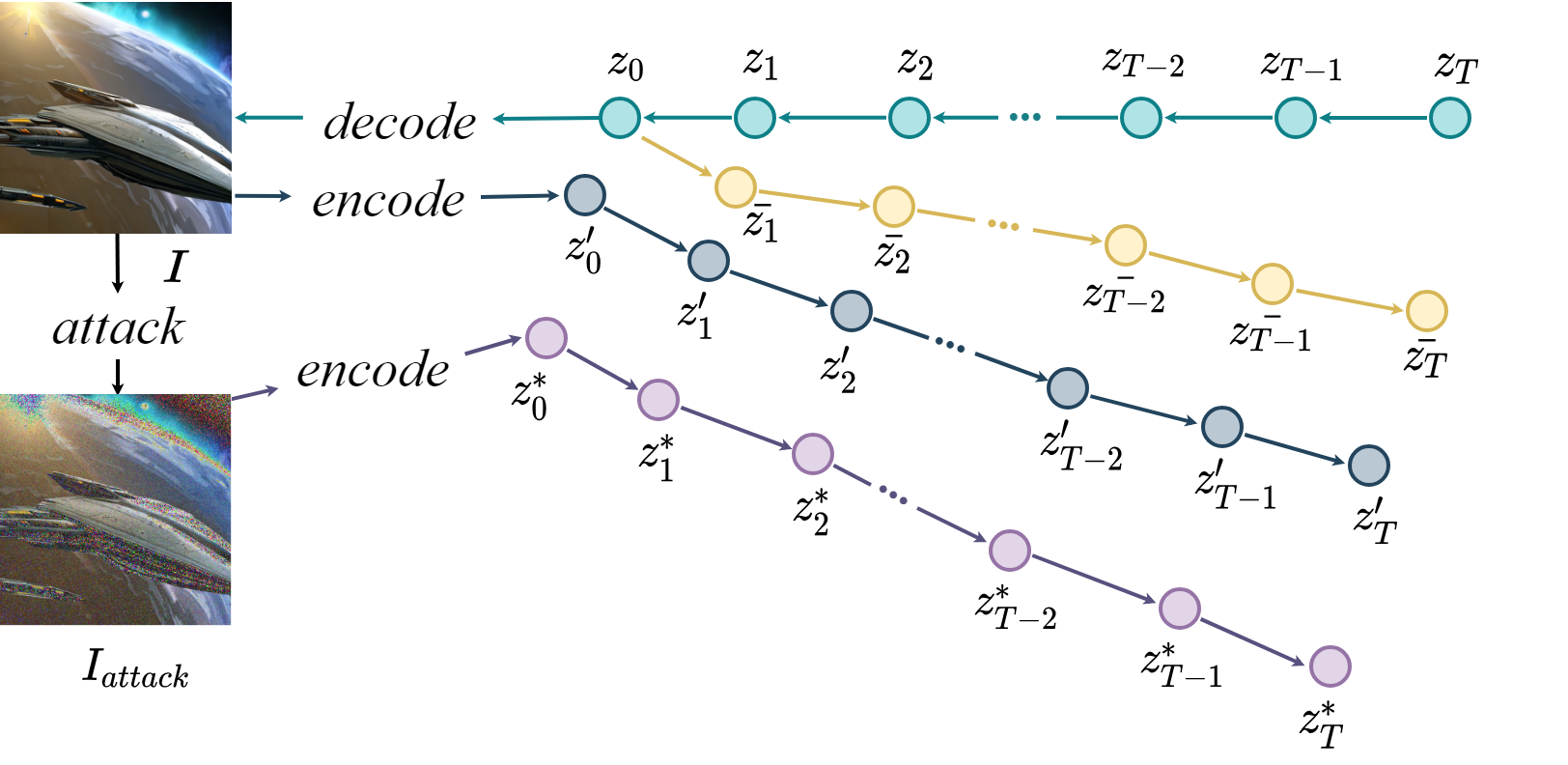}
    }
    \subfloat[\label{fig:inversion_error}]{
        \includegraphics[width=0.32\textwidth]{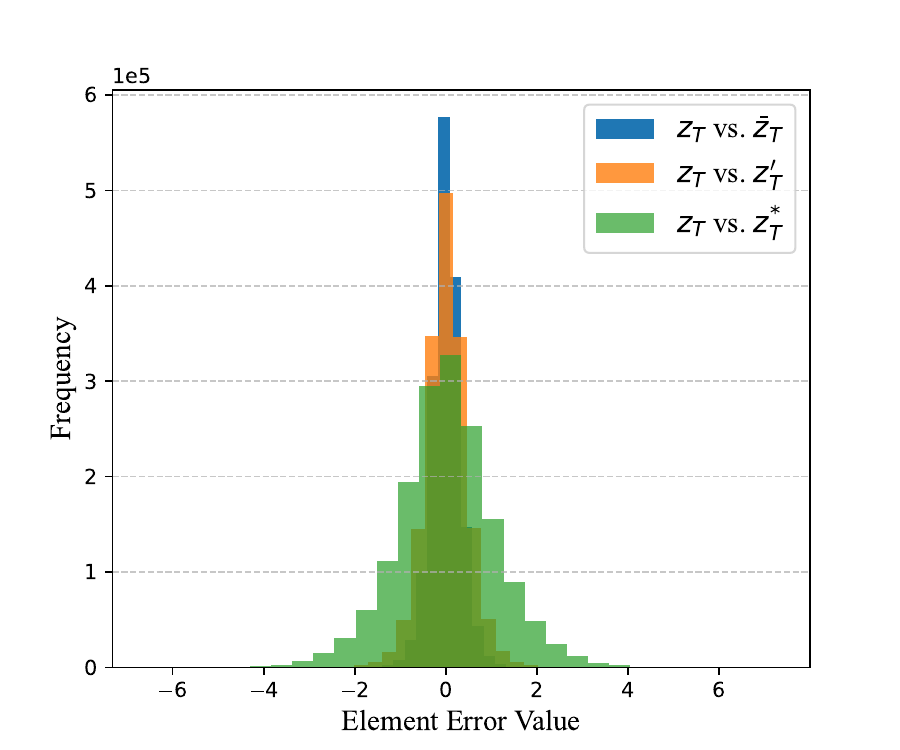}
    }
    \subfloat[\label{fig:sign_consistency}]{
        \includegraphics[width=0.32\textwidth]{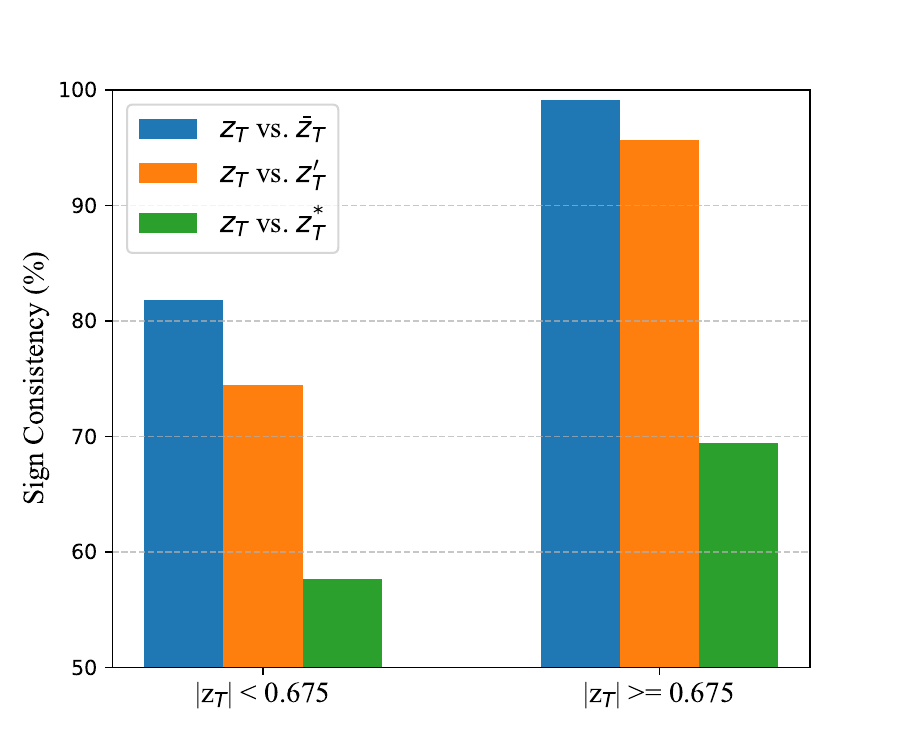}
    }

    \caption{
        Illustration of the denoising and inversion process and analysis of inversion errors. 
        (a) Visualizes the denoising and inversion process in the latent diffusion model (LDM). The original image $I$ is encoded into the latent space, and the latent variable $z_0$ is recovered through inversion, while the attacked image $I_{\text{attack}}$ introduces further discrepancies in the recovered latent variables. 
        (b) Presents the distribution of inversion errors under different conditions, showing the error values between $z_T$ and its reconstructed counterparts ($\bar{z}_T$, $z_T'$, and $z_T^*$). 
        (c) Evaluates the sign consistency of latent elements, highlighting that elements with larger absolute values ($|z_T| \geq 0.675$) exhibit higher robustness to inversion errors compared to smaller absolute values ($|z_T| < 0.675$).
    }
    \label{fig:denoising_inversion_analysis}
\end{figure*}

\subsection{Detection Watermarking}

Detection watermarking is used to verify whether an image is generated by the service provider’s model. The service provider embeds a unique detection watermark into each generated image to distinguish it from natural images or images generated by other models. Assume that \( m \) represents the watermark embedded in the generated image and \( m' \) represents the watermark extracted from a test image. If the number of matching bits between \( m \) and \( m' \) exceeds a predefined threshold \( \tau \), which corresponds to the number of correct watermark bits required for a specified false-positive rate, the image is classified as being generated by the service provider’s model. The detection process must be robust against noise addition, image transformations, and both intentional and unintentional modifications.

\subsection{Attribution Watermarking}

Attribution watermarking enables content traceability by linking generated images to their originating users. In cases where malicious users attempt to bypass watermarking mechanisms to generate illicit content, attribution watermarking can identify the responsible user, thereby assigning accountability. For scenarios involving the tampering of other users’ content, attribution watermarking can trace the image back to its original user, providing evidence for copyright claims. However, malicious users may claim that their images were tampered with by others, complicating the judgment process. To address this, the service provider can record contextual information, such as the generation timestamp and input descriptions. Combined with attribution watermarking, this information can assist in determining the intent and legitimacy of the generation process. 

Therefore, attribution watermarking extends detection watermarking by embedding a unique watermark for each user, enabling precise traceability of generated images. This mechanism not only verifies whether an image was generated by the service provider’s model but also identifies the specific user responsible for the generation. The attribution process involves comparing the extracted watermark \( m' \) with a set of user-specific signatures \(\{m^{(1)}, m^{(2)}, \dots, m^{(N)}\}\), where \( N \) represents the total number of registered users. If the number of correct bits in the extracted watermark matches a user’s signature beyond the threshold \( \tau \), the service provider can attribute the generated image to the corresponding user, ensuring accountability and copyright protection. As the number of users increases, the number of correct watermark bits required for attribution is generally higher than for detection watermarking to maintain an acceptable false-positive rate.

\section{Proposed Method}
\subsection{Preliminary Analysis}

\begin{figure*}[htbp]
    \centering
    \includegraphics[width=\textwidth]{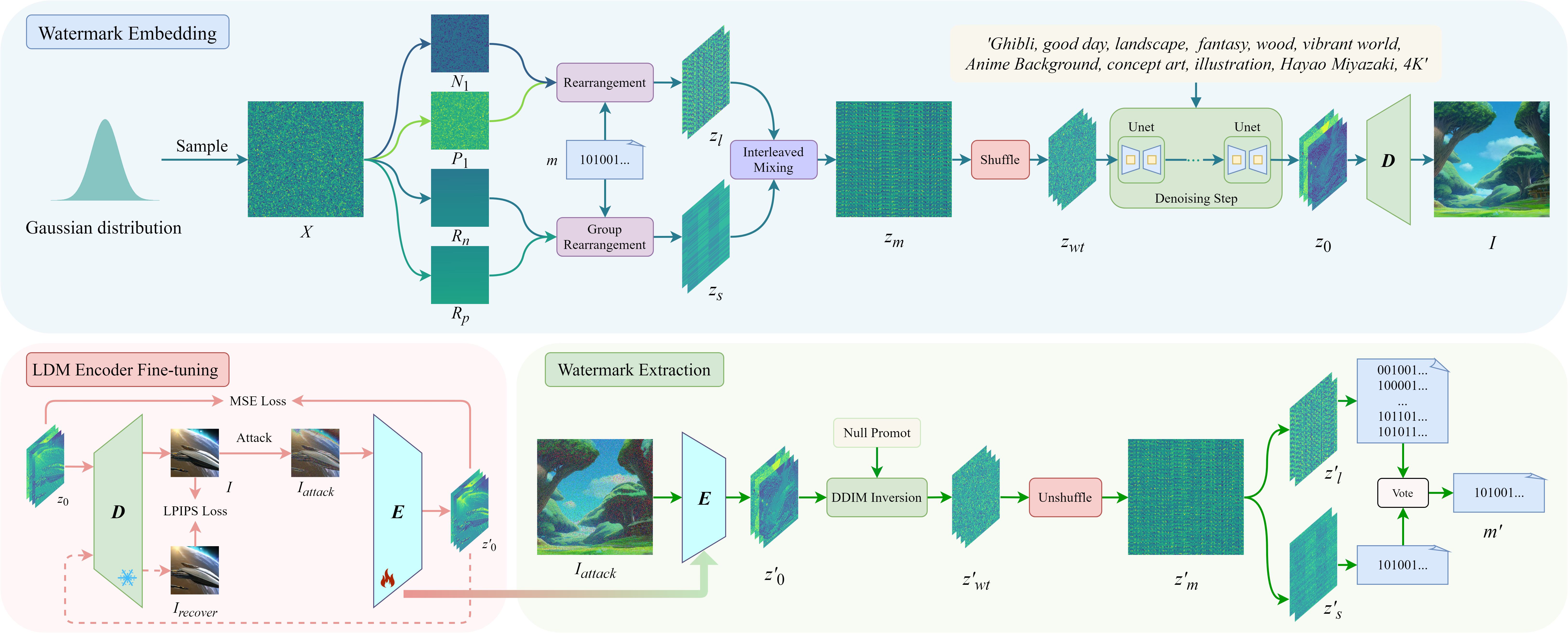}
    \caption{The framework of TraceMark-LDM. The watermark information \(m\) is used to guide a rearrangement process to produce a latent input \(z_{wt}\) for generating watermarked image \(I\). During the fine-tuning phase, the decoder parameters are frozen and only the encoder parameters are updated. The extraction process is an inversion of the embedding. } 
    \label{fig:framework}
\end{figure*}

To develop a robust and lossless watermark embedding strategy, we conduct a systematic analysis of the denoising and inversion processes inherent in diffusion models, yields the following key observations:

1. \textbf{Minimizing the difference in latent representation of distorted images helps reduce inversion errors.} 

As shown in Fig. \ref{fig:denoising_inversion_process}, the denoising and inversion process of the diffusion model is demonstrated. When the denoised latent variable \( z_0 \) is decoded to produce the generated image \( I \), and then re-encoded back into latent space, the resulting latent representation \( z_0' \) exhibits a difference from the original \( z_0 \). This indicates that the VAE encoding-decoding process significantly affects the image's latent representation. Recent research on numerical errors in diffusion ODE inversion \cite{hu2024establishing} has pointed out that the bidirectional mapping process from \( z_0 \) to \( z_t \) in the latent space is highly fragile, and even minor differences can accumulate during the inversion process. This suggests that VAE encoding-decoding operations increase inversion errors. Furthermore, when the image is subjected to an attack, the difference between the recovered noise after inversion and the original initial noise is greater compared to the recovery from an unperturbed image.

In Fig. \ref{fig:inversion_error}, we quantify these errors under different conditions. The results show that both VAE encoding-decoding operations and image distortions increase the inversion error. However, when the latent variable \( z_0 \) is not subjected to encoding-decoding, the inversion recovery from \( z_T \) to the initial \( \bar{z}_T \) has the smallest difference, with errors concentrated in the range of [-0.6, 0.6], symmetrically distributed around zero. This indicates that minimizing the latent variable \( z_0^* \) of the distorted image with \( z_0 \) can effectively reduce the impact of distortion and encoding-decoding operations on inversion errors.

2. \textbf{The sign after denoising and inversion of large absolute value elements have high stability.}  

Based on the analysis of inversion errors, we classified all elements of the initial noise according to the quartiles of the standard normal distribution, dividing them into large absolute value elements (\(|z_T| \geq 0.675\)) and small absolute value elements (\(|z_T| < 0.675\)). Then statistically the consistency of their element sign after denoising and inversion. As shown in Fig. \ref{fig:sign_consistency}, the large absolute value elements maintain over 95\% sign consistency after inversion, and even under distorted conditions, this consistency remains close to 70\%. This characteristic allows them to be effectively used for marking watermark information. In contrast, elements with small absolute values exhibit greater instability and lower sign consistency, making them unsuitable for marking watermarking information with a single element.

3. \textbf{Distributional invariance of i.i.d. Gaussian samples under random permutations.} Consider a random vector \( X = (X_1,\dots,X_n) \) whose components are independent and identically distributed (i.i.d.) standard Gaussian random variables, 
\(
X_i \sim \mathcal{N}(0,1) \quad \text{for } i=1,2,\dots,n,
\)
with each \(X_i\) independent of the others. Hence, the joint probability density function (pdf) is
\begin{equation}
f_X(X_1,\dots,X_n) = \prod_{i=1}^n \frac{1}{\sqrt{2\pi}} \exp\left(-\frac{X_i^2}{2}\right)
\label{eq:fx_pdf}
\end{equation}

Let \(\mathcal{P}\) be a random permutation of \(\{1,\dots,n\}\) chosen uniformly at random and independent of \(\{X_i\}\):
\begin{equation}
P(\mathcal{P}=\sigma) = \frac{1}{n!} \quad \forall \sigma \in S_n,
\label{eq:p}
\end{equation}
where \(S_n\) denotes the symmetric group on \(n\) elements.

Applying \(\mathcal{P}\) to \(X\) yields the permuted vector \(X'\).

By the i.i.d. and hence exchangeable nature of the Gaussian components, for any fixed \(\sigma \in S_n\),
\begin{equation}
(X_{\sigma(1)}, \dots, X_{\sigma(n)}) \stackrel{d}{=} (X_1,\dots,X_n).
\label{eq:same_distribution_1}
\end{equation}

Conditioning on \(\mathcal{P}\) and substituting \(P(\mathcal{P}=\sigma)=\tfrac{1}{n!}\) gives
\begin{equation}
\begin{aligned}
f_{X'}(X_1,\dots,X_n) &= \sum_{\sigma \in S_n} f_{X'|\mathcal{P}=\sigma}(X_1,\dots,X_n) P(\mathcal{P}=\sigma) \\
&= \sum_{\sigma \in S_n} \left(\prod_{i=1}^n \frac{1}{\sqrt{2\pi}}\exp\left(-\frac{X_i^2}{2}\right)\right)\frac{1}{n!} \\
&= \prod_{i=1}^n \frac{1}{\sqrt{2\pi}}\exp\left(-\frac{X_i^2}{2}\right).
\label{eq:fx'_1}
\end{aligned}
\end{equation}

In conclusion, after applying a uniformly random, independent permutation, the resulting vector \(X'\) retains the original distribution of \(X\), i.e., \(X' \stackrel{d}{=} X\).

Based on the sign consistency and distributional invariance mentioned above, we propose a novel watermarking algorithm for LDM, as shown in Fig. \ref{fig:framework}. The procedure is structured into three distinct phases: watermark embedding, LDM encoder fine-tuning, and watermark extraction. During the embedding stage of the watermark, a random variable sampled from a Gaussian distribution is divided into four subsets according to the element values. These subsets are later rearranged according to the watermark information and then fed into the generative model to generate content. The purpose of fine-tuning the LDM encoder stage is to minimize errors caused by attacks or disturbances on the latent representation of generated images. During the watermark extraction stage, the generated images are reversed back to their initial latent noise domain via DDIM inversion. This process enables the extraction of watermark information based on the pre-established arrangement. The following sections will provide a comprehensive explanation and validation of this method's implementation.

\subsection{Watermark Embedding}
To maintain the high quality of the generated images, we sample a random variable \( X = \{ x_1, x_2, x_3, \ldots, x_{(r-1)}, x_r \} \) containing \( r \) elements following a Gaussian distribution. Here, \( r = c \times h \times w \), where \( c \), \( h \), and \( w \) are the dimensions of the latent space in the LDM. The watermark \( m \) is a random binary bitstream of \( k \) bits, equally distributed between 0 and 1.

The core concept of our watermark embedding approach leverages the inherent properties of noise elements in diffusion models, specifically utilizing their positive and negative values to encode watermark information. Therefore, the random variables \( X \) are divided into a negative set \( N \) and a non-negative set \( P \):
\begin{equation}
\begin{aligned}
 N = \{ x_t \in X \mid x_t < 0 \} \\
 P = \{ x_t \in X \mid x_t \geq 0 \}
\end{aligned}
\label{eq:4}
\end{equation}

\subsubsection{Binary Guided Rearrangement}
According to the previous analysis, elements with large absolute values are less susceptible to denoising and inversion, and a single element can be used to indicate watermarking information. Based on the division of standard normal distribution quartiles, we select larger absolute value elements from the sets \( N \) and \( P \), with each set accounting for a quarter of the total number of elements. These are denoted as \( N_1 \) and \( P_1 \), respectively.

\begin{equation}
N_1 = \{n \in N \mid \textit{rank}(n, |N|) \leq \frac{r}{4}\}
\label{eq:N1}
\end{equation}
\begin{equation}
P_1 = \{p \in P \mid \textit{rank}(p, |P|) \leq \frac{r}{4}\}
\label{eq:P1}
\end{equation}

Here, \(\textit{rank}(x, S)\) represents the rank of \(x\) within the set \(S\) when sorted by absolute value in descending order.

The watermark information \( m \) directs the rearrangement of elements within the \( N_1 \) and \( P_1 \) sets. Specifically, if the watermark bit is 0, an element from \( N_1 \) is selected; If the watermark bit is 1, an element from \( P_1 \) is chosen. To enhance the fault tolerance of the watermark, \( m \) is repeatedly used, resulting in a sequence \( z_l \) consisting of elements with larger absolute values.

\begin{algorithm}[t]
\caption{Watermark Embedding Process}
\label{alg:embedding_process}
     \textbf{Input:} Watermark bitstream \(m\) of length \(k\) bits, and the model key \(s\).\\
     \textbf{Output:} Watermarked generated image \( I \)

\begin{algorithmic}[1]
    \STATE Obtain the Gaussian variable \( X \in \mathbb{R}^{r} \) by randomly sampling each component from a standard Gaussian distribution.
    \STATE Partition \( X \) into \( P \) and \( N \) based on their positive and negative values.
    \STATE Order the elements of \(N\) by their absolute values and form \(N_1\) by selecting those in the top quarter.
    \STATE Similarly, order the elements of \(P\) by their absolute values and form \(P_1\) by selecting those in the top quarter.
    \STATE Initialize the empty sets \( z_l \) and \( z_s \).
    \FOR{each \( t \in [1,\frac{r}{2k} ]\)}
        \FOR {each \( b_i  \in  m \)}
            \STATE According to the value of \( b_i \), select the corresponding element from \( N_1 \) or \( P_1 \) and add it to the set \( z_l \).
        \ENDFOR
    \ENDFOR
    \STATE Define \( R \) as the union of all elements in \( P \) that are not in \( P_1 \) and all elements in \( N \) that are not in \( N_1 \).
    \STATE Sort \( R \), then divide it equally into two parts: \( R_n \) and \( R_p \).
    \STATE Obtain \( G_n \) and \( G_p \) by grouping \( R_n \) and \( R_p \), respectively.

    \FOR {each bit \( b_i \in m \)}
        \STATE According to the value of \( b_i \), select the corresponding group from \( G_n \) or \( G_p \), and add all elements of the selected group to the set \( z_s \).
    \ENDFOR
    
    \STATE Interleave \( z_l \) and \( z_s \) to obtain \( z_m \).
    \STATE Shuffle the positions of \( z_m \) using the assigned model key \( s \) to obtain \( z_{wt} \).
    \STATE Obtain the latent variable \( z_0 \) by applying multi-step denoising to \( z_{wt} \).
    \STATE Decode \( z_0 \) using the LDM decoder to generate the watermarked image \( I \).

    \RETURN \( I \)

\end{algorithmic}
\end{algorithm}

\subsubsection{Binary Guided Group Rearrangement}

As demonstrated in Fig. \ref{fig:sign_consistency}, in the case of image distortion, only approximately 55\% of the elements with small absolute values retain their original sign after the inversion process. This considerable instability suggests that relying on the sign of an individual element for watermarking significantly compromises the robustness of the method. Consequently, a more effective approach is to avoid using such elements in the embedding process.

To fully exploit the potential of these small absolute value elements, we introduce a group rearrangement strategy. Instead of treating them individually, we logically organize them into groups, ensuring that the watermark information is conveyed through the sign of the sum of each group. This strategy enhances the authentication performance by mitigating the impact of sign flipping at the individual element level. The specific operations are as follows:

The remaining half of elements in \(X\) with smaller absolute values are combined to form set \(R\), containing \(\frac{r}{2}\) elements, defined as:
\begin{equation}
R = \{ x_i \in P | x_i \notin P_1 \} \cup \{ x_i \in N | x_i \notin N_1 \}
\label{eq:R}
\end{equation}

Due to stochastic sampling, the numbers of positive and negative elements in \(R\) may not be strictly balanced. To mitigate this imbalance, \(R\) is sorted in ascending order and equally divided into two subsets \(R_n\) and \(R_p\), each containing \(\frac{r}{4}\) elements:
\begin{equation}
R_n = \{R_{n,i} \mid R_{n,i} \in \textit{sorted}(R)[1, \frac{r}{4}]\}
\label{eq:Rn}
\end{equation}
\begin{equation}
 R_p = \{R_{p,i} \mid R_{p,i} \in \textit{sorted}(R)[\frac{r}{4}+1, \frac{r}{2}]\}
\label{eq:Rp}
\end{equation}

Subsequently, \(R_n\) and \(R_p\) are partitioned into subsets \(G_n\) and \(G_p\) via a symmetric grouping strategy. Taking \(R_n\) as an example, this ordered sequence is divided into \(\frac{k}{2}\) groups by alternately selecting elements from both ends and dynamically allocating them to minimize inter-group sum differences, forming subset \(G_n\). Similarly, \(R_p\) is processed to generate \(G_p\). Formally:
\begin{equation} 
\begin{aligned}
G_n = \{ G_{n,1}, G_{n,2}, G_{n,3}, \ldots, G_{n,\frac{k}{2}} \} \\
G_p = \{ G_{p,1}, G_{p,2}, G_{p,3}, \ldots, G_{p,\frac{k}{2}} \}
\end{aligned}
\label{eq:9}
\end{equation}

Compared to using individual small absolute elements' signs for watermark encoding, the summed sign of grouped elements exhibits enhanced robustness. For instance, a group in \(G_n\) containing multiple negative elements has an initial sum \(v\) encoding watermark bit 0. During extraction, although element values may deviate due to inversion errors (as shown in Fig. \ref{fig:inversion_error}, where errors follow a zero-mean symmetric distribution), positive and negative errors partially cancel during summation. Since \(v\) is derived from cumulative contributions of elements of the same sign (e.g., a strongly negative sum for negative groups), its sign demonstrates higher resilience to error interference.

Leveraging this symmetry-driven optimization, watermark embedding proceeds via group-wise rearrangement: if a watermark bit is 0, an unused group from \(G_n\) is selected; if 1, a group from \(G_p\) is chosen. Selected groups are sequentially concatenated to form the encoded sequence \(z_{\text{s}}\). The resulting sequence \(z_{\text{s}}\) is a rearrangement of R that both preserves the distributional properties of the elements and implicitly carries watermarking information through the sequence of sign of the group sum.

\subsubsection{Interleaved Mixing and Shuffle}
The \(r\) elements from \(z_{\textit{l}}\) and \(z_{\textit{s}}\) are integrated using an interleaved mixing method to form \(z_{\textit{m}}\), which initially disrupts ordered features. In this method, elements from \( z_l \) and \( z_s \) are alternately selected to create \(z_{\textit{m}}\). For example, if \( z_l \) = \{-1.2, 1.2, 1.1, -1.1, -1.0, 1.0, 0.9, -0.9\} and \( z_s \) = \{-0.4, -0.1, 0.4, 0.1, 0.3, 0.2, -0.3, -0.2\}, and the watermark is 0110, the process starts by selecting four elements from \( z_s \), followed by four elements from \( z_l \), alternating in this pattern. The resulting \( z_m \) = \{-1.2, 1.2, 1.1, -1.1, -0.4, -0.1, 0.4, 0.1, -1.0, 1.0, 0.9, -0.9, 0.3, 0.2, -0.3, -0.2\}. \(z_{\textit{m}}\) is then shuffled and reshaped using the model key \( s \), producing the initial latent noise \(z_{\textit{wt}}\).  

\subsubsection{Image Generation}
After denoising steps, the latent variable \(z_0\) for generating the image is obtained. It is subsequently mapped to the spatial domain through the LDM decoder, generating the image \(I = D(z_0)\). The detailed steps of the embedding algorithm are provided in Algorithm \ref{alg:embedding_process}.

\subsection{LDM Encoder Fine-Tuning}

Initial noise sampling watermarking methods rely on DDIM to invert generated images back to the initial noise. However, the inversion errors introduced in this process significantly degrade the robustness of watermarks. To address this issue, we propose fine-tuning the LDM encoder to adapt to various complex distortion scenarios, thereby effectively reducing inversion errors. Since the LDM encoder maps images from pixel space back to latent space without participating in the generative process, the fine-tuning process does not affect the quality of generated images. After fine-tuning, the encoder ensures that the latent variables of distorted images are closer to the original latent variables.

As mentioned earlier, directly inverting the original latent variables results in the smallest error in the initial noise. This suggests that minimizing the differences between latent variables can effectively reduce the inversion error of the initial noise. 

Initially, the image \(I = D(z_0)\) is generated by decoding the original latent variable \(z_0\) using the LDM decoder \(D\).
Subsequently, random perturbations are applied to the image \(I\) to simulate a distortion scenario, resulting in the perturbed image \(I'\).

Next, the perturbed image \(I'\) is fed into the fine-tuned LDM encoder \(E\), which maps it back into the latent space to obtain the latent variable \(\hat{z}_0 = E(I')\). The mean squared error (MSE) loss between \(z_0\) and \(\hat{z}_0\) is calculated as follows:
\begin{equation}
\mathcal{L}_{\textit{inv}} = \min_{\theta} \mathbb{E} \left[ \| z_0 - E_{\theta}(I') \|^2 \right]
\label{eq:1}
\end{equation}

By optimizing the MSE loss, we minimize the difference between the latent variables of the generated image before and after perturbation, thereby reducing the inversion error of the initial noise.

Further, \(\hat{z}_0\) is fed into the decoder \(D\), which converts it back to the spatial domain to synthesize the recovered image \(I_{\textit{recover}} = D(\hat{z}_0)\). The perceptual similarity (LPIPS) loss between the original image \(I\) and the recovered image \(I_{\textit{recover}}\) is then calculated as follows:
\begin{equation}
\mathcal{L}_{\textit{sim}} = \min_{\theta} \mathbb{E} \left[ \textit{LPIPS}(D(z_0), D(E_{\theta}(I'))) \right]
\label{eq:2}
\end{equation}

Finally, the total loss function is optimized to balance two loss terms. This ensures that the fine-tuning minimizes differences between initial noise \(z_0\) and \(\hat{z}_0\) under various perturbations, without introducing additional noise during the inversion process of the diffusion model:
\begin{equation}
 \mathcal{L} = \mathcal{L}_{\textit{inv}} + \lambda \mathcal{L}_{\textit{sim}}
\label{eq:3}
\end{equation}
where \(\lambda\) is a weight coefficient used to balance the two loss terms. It is worth noting that the fine-tuning process does not require the diffusion model's inversion steps. This not only significantly improves the efficiency of fine-tuning but also greatly reduces computational costs. 

\begin{algorithm}[t]
\caption{Watermark Extraction Process}
\label{alg:extraction_process}
     \textbf{Input:} Generated image \( I \),  and the model key \( s \). \\
     \textbf{Output:} Extracted watermark \( m' \)
     
\begin{algorithmic}[1]
    \STATE Encode \( I \) using the LDM encoder to obtain the latent variable \( z_0' \).
    \STATE Obtain \( z_{wt} \) by performing multi-step inversion on \( z_0 \).
    \STATE Unshuffle the positions of \( z_{wt}' \) using the assigned model key \( s \) to obtain \( z_m' \)
    \STATE Extract \( z_l \) and \( z_s \) by deinterleaving \( z_m \).

    \FOR {each item \( p  \in  z_l' \)}
        \STATE Determine the watermark value as 0 or 1 based on the sign of \( p \).
    \ENDFOR
    
    \FOR {each group \( g  \in  z_s' \)}
        \STATE Determine the watermark value as 0 or 1 based on the sign of the sum of elements within group \( p \).

    \ENDFOR
    
    \STATE Aggregate multiple watermark groups through voting to obtain the final watermark \( m' \).
    \RETURN \( m' \)
\end{algorithmic}
\end{algorithm}
\subsection{Watermark Extraction Process}
The watermark extraction process is essentially the reverse of the embedding process. We begin by feeding the generated image into the fine-tuned LDM encoder to retrieve the latent variable \( z'_{0} \), denoted as \( z'_{0} = E'(I)\). Next, the initial watermarked noise \( z'_{\text{wt}} \) is recovered through DDIM inversion. The unshuffle operation using the model key \(s\) in the embedding process is then applied to \( z'_{\text{wt}} \) to obtain the interleaved noise \( z'_{\text{m}} \). This conversion aims to reverse the random permutation applied during the embedding process. Subsequently, we perform the deinterleave operation on \( z'_{\text{m}} \), separating it into two parts: \( z'_{\text{l}} \) and \( z'_{\text{s}} \).

For each element in \( z'_{l} \), the watermark bit is determined by its sign: negative elements represent a watermark bit of 0, while non-negative elements represent a bit of 1. This process yields the watermark bitstream \( w_1 \). Similarly, for each group of elements in \( z'_{s} \), the sum of each group is computed to determine the watermark bit, producing the watermark bitstream \( w_2 \). Next, \( w_1 \) and \( w_2 \) are merged to form a complete bitstream \( w \).

Then, \(  w \) is divided into multiple substreams, each with a length equal to that of the original watermark. Finally, a voting mechanism is employed to determine the recovered watermark bitstream \( m' \). The detailed steps of the extraction algorithm are provided in Algorithm \ref{alg:extraction_process}.

\begin{figure*}[ht]
\centering
	\subfloat[\label{fig:median_filter}]{\includegraphics[width = 0.24\textwidth]{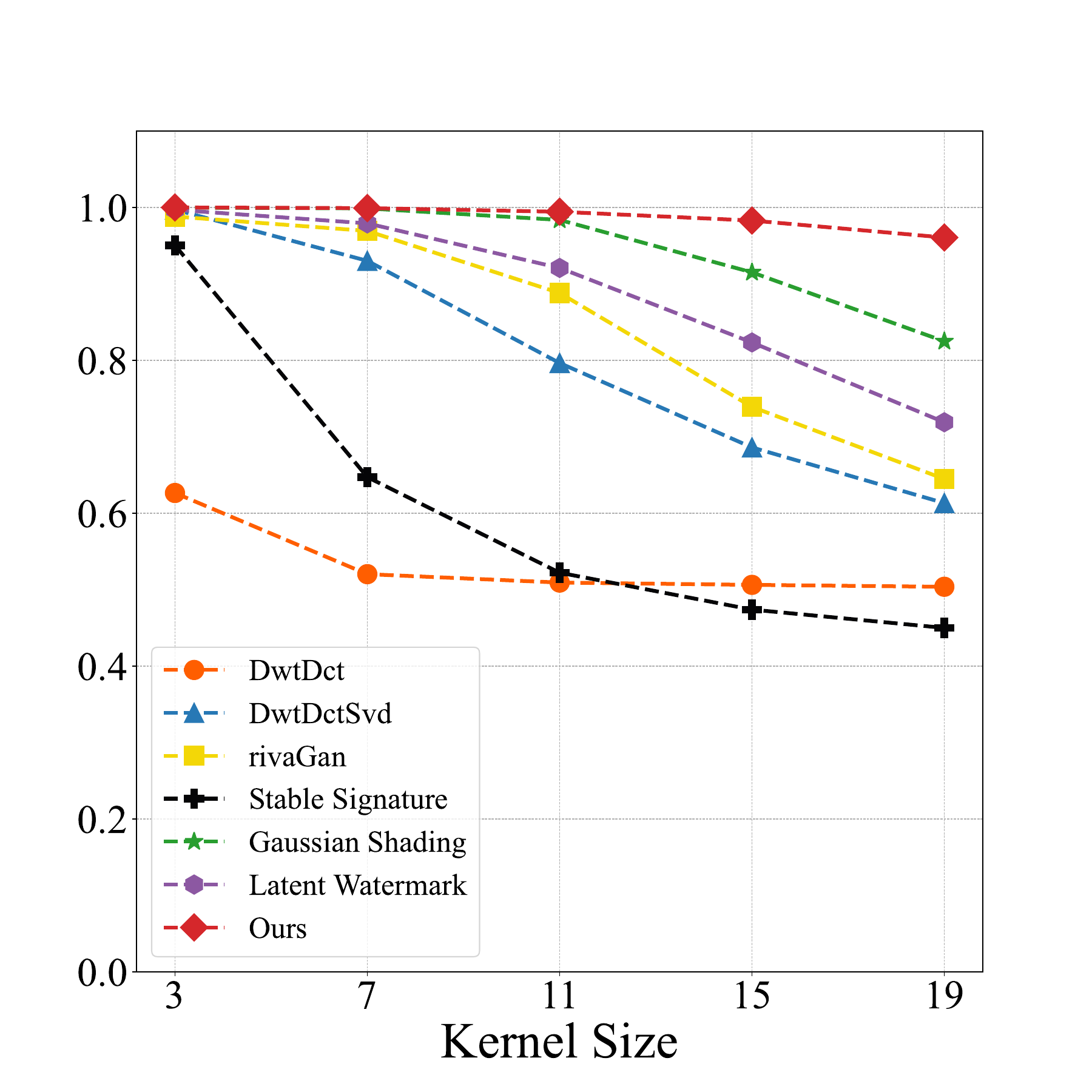}}
	% \hfill
	\subfloat[\label{fig:jpeg_compression}]{\includegraphics[width = 0.24\textwidth]{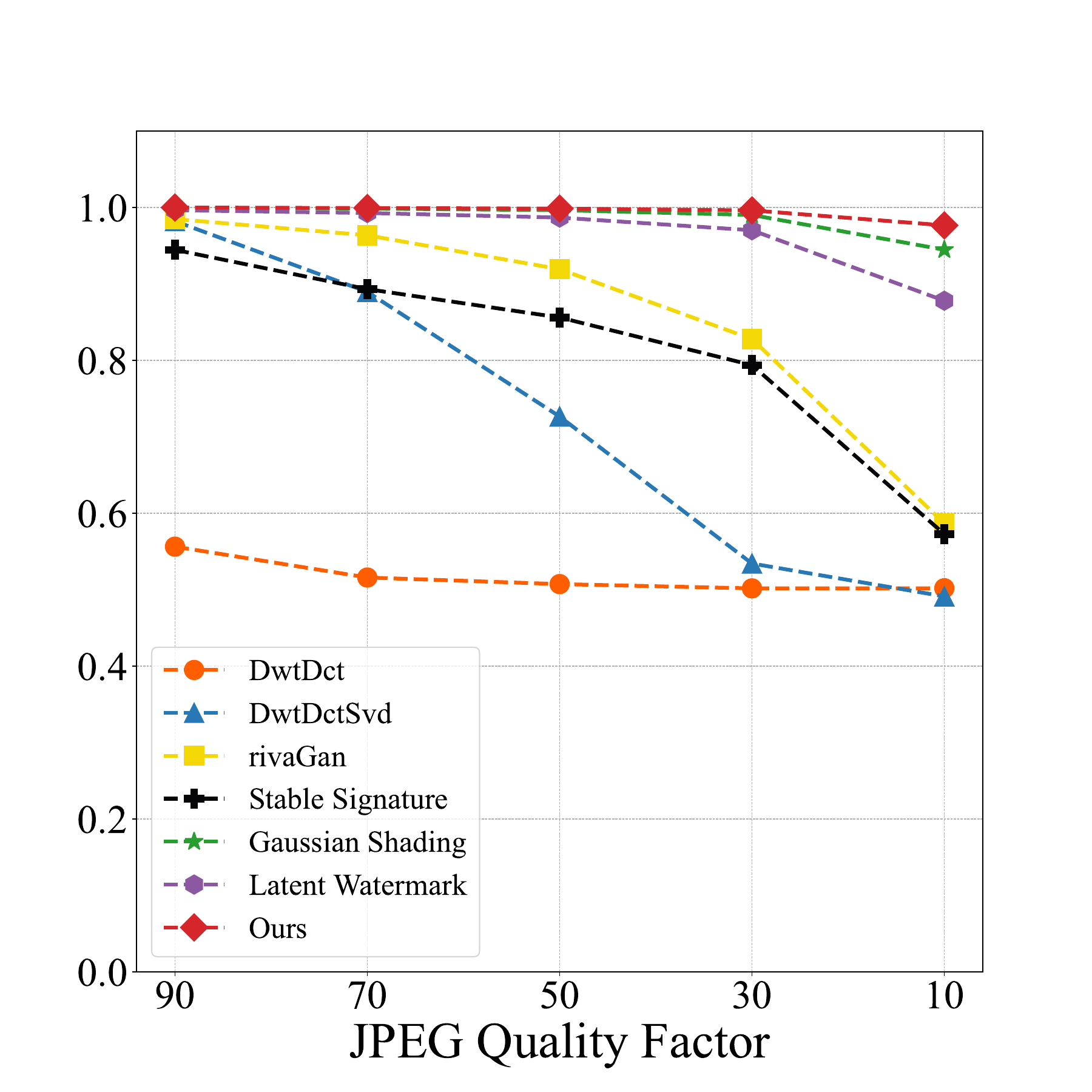}}
    % \hfill
	\subfloat[\label{fig:gaussian_blur}]{\includegraphics[width = 0.24\textwidth]{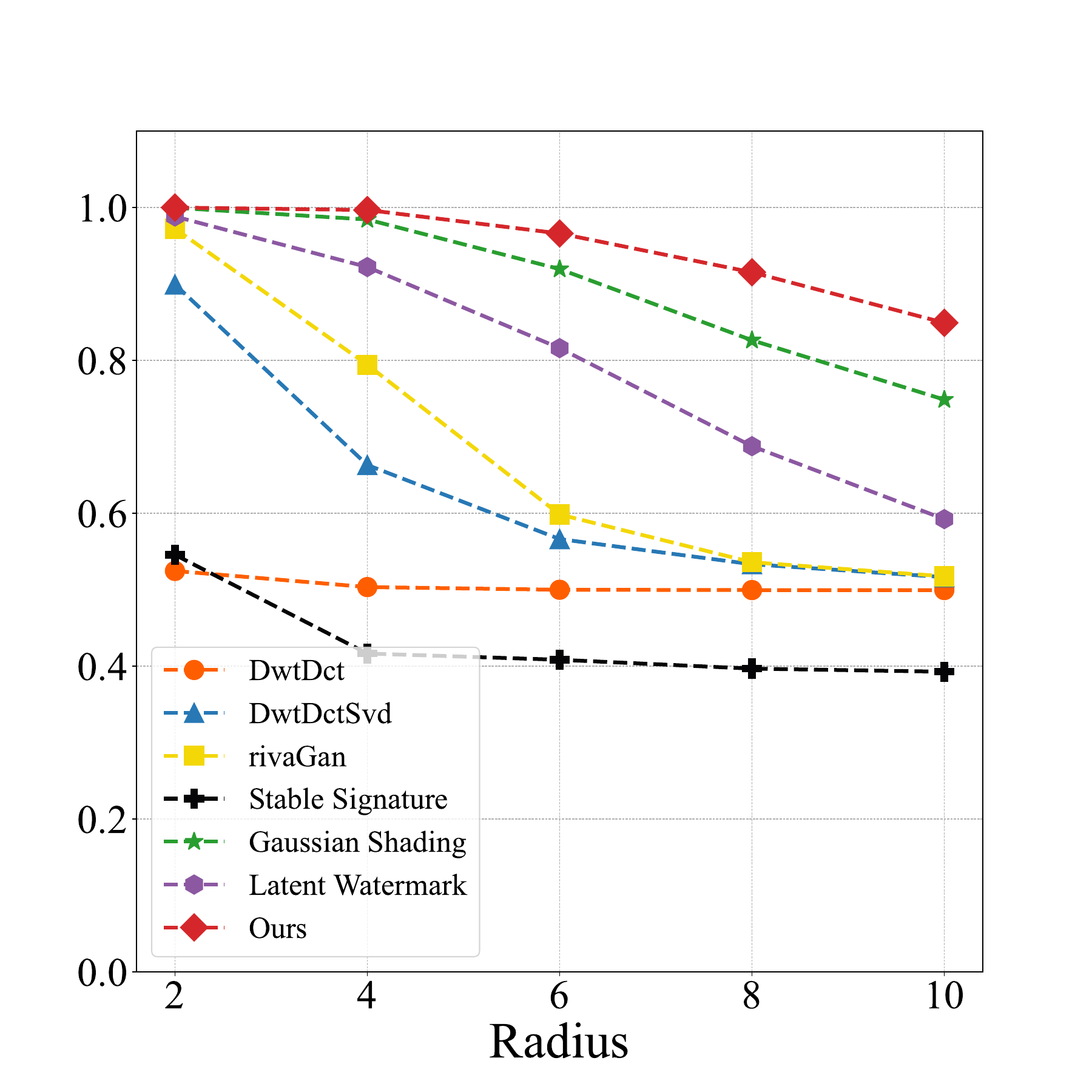}}
    % \hfill
	\subfloat[\label{fig:gaussian_noise}]{\includegraphics[width = 0.24\textwidth]{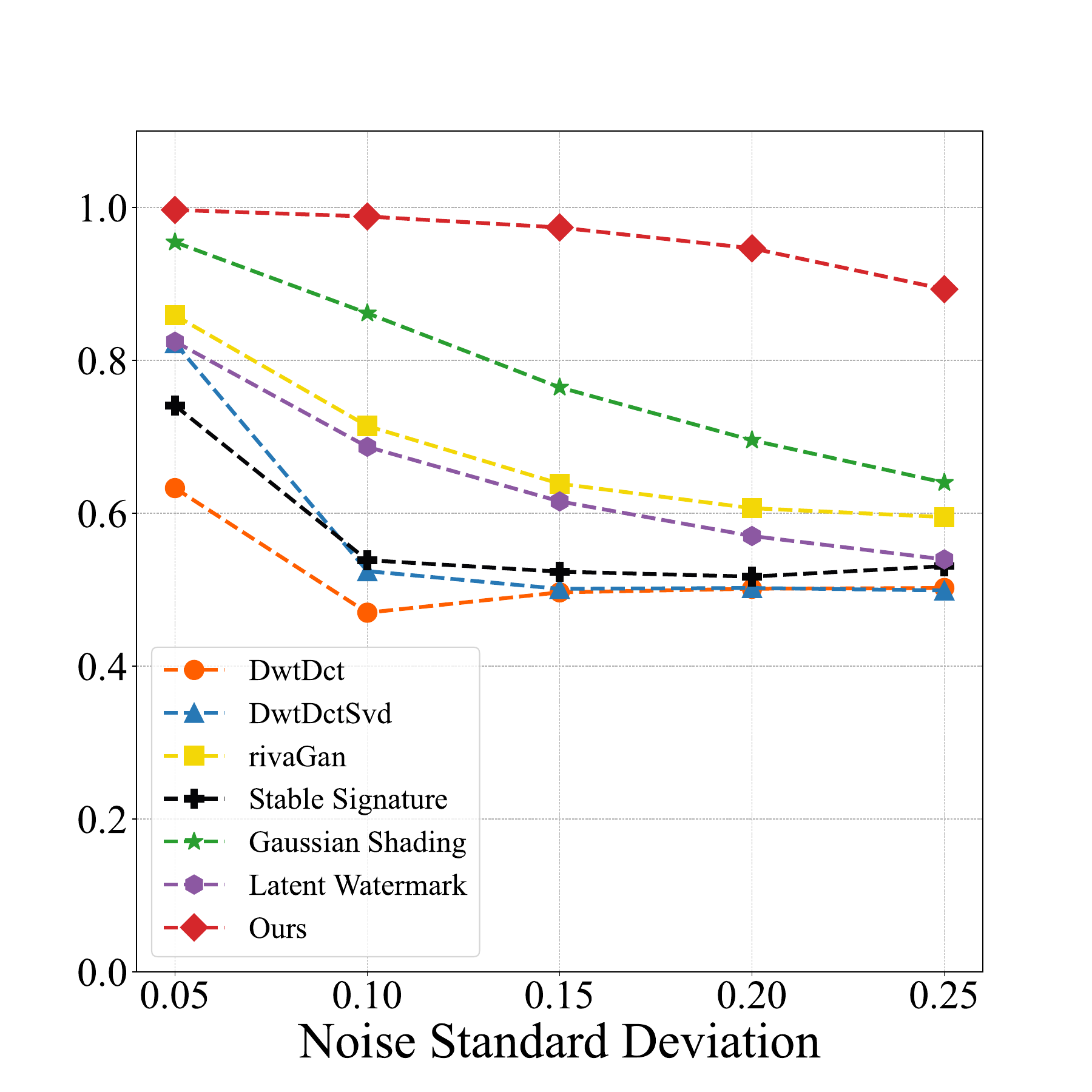}}
    \hfill
	\subfloat[\label{fig:sp_noise}]{\includegraphics[width = 0.24\textwidth]{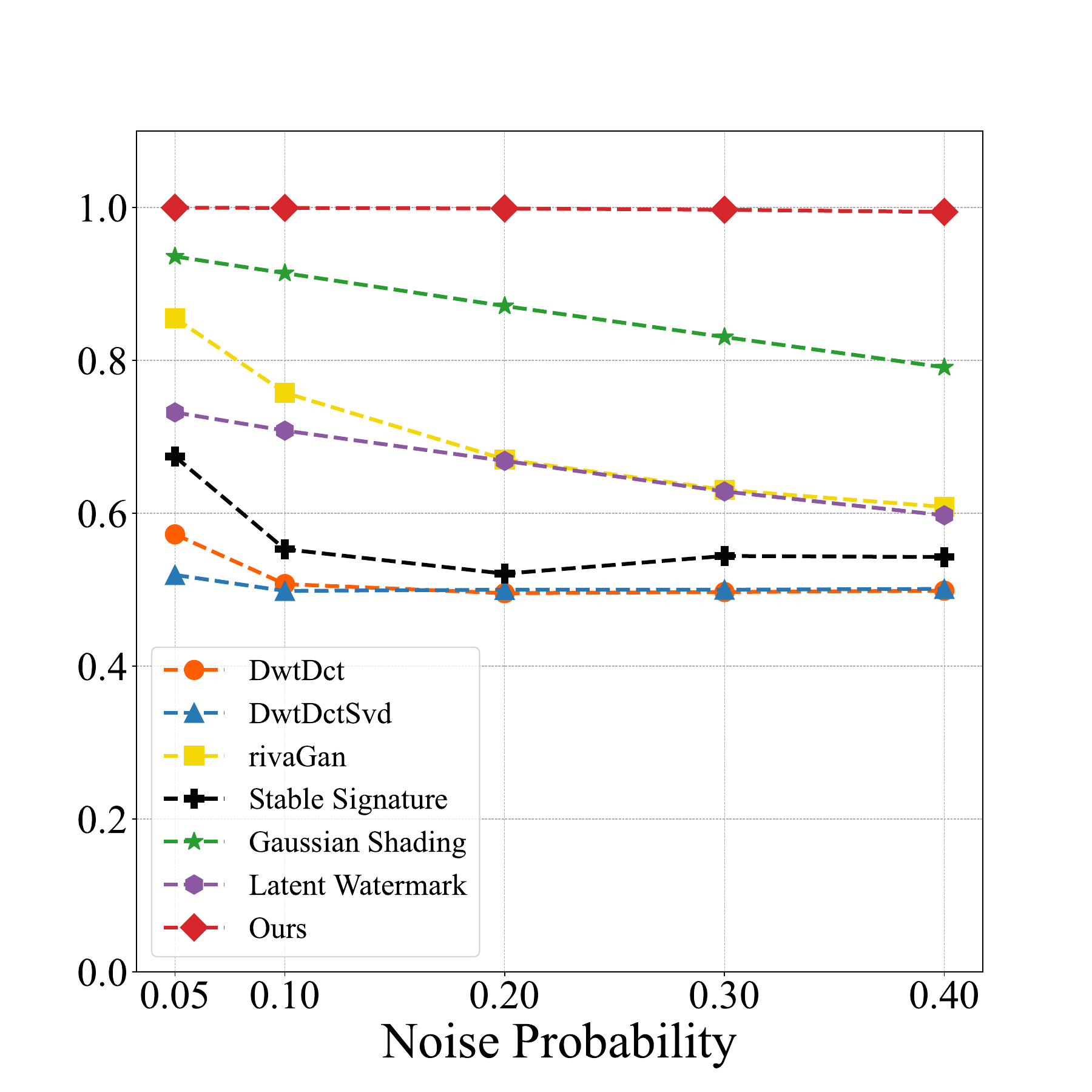}}
    % \hfill
	\subfloat[\label{fig:resize_restore}]{\includegraphics[width = 0.24\textwidth]{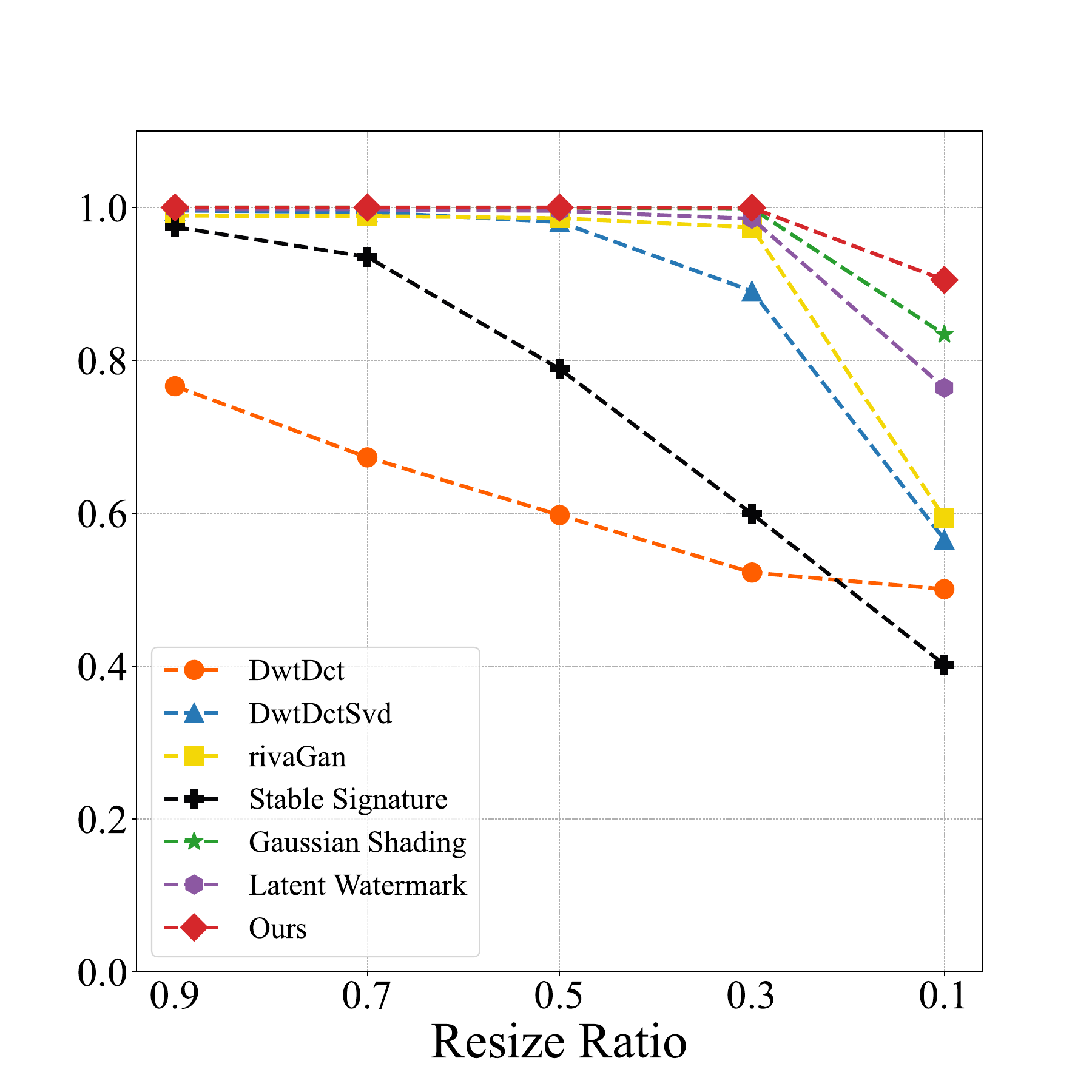}}
    % \hfill
	\subfloat[\label{fig:bmshj2018_hyperprior}]{\includegraphics[width = 0.24\textwidth]{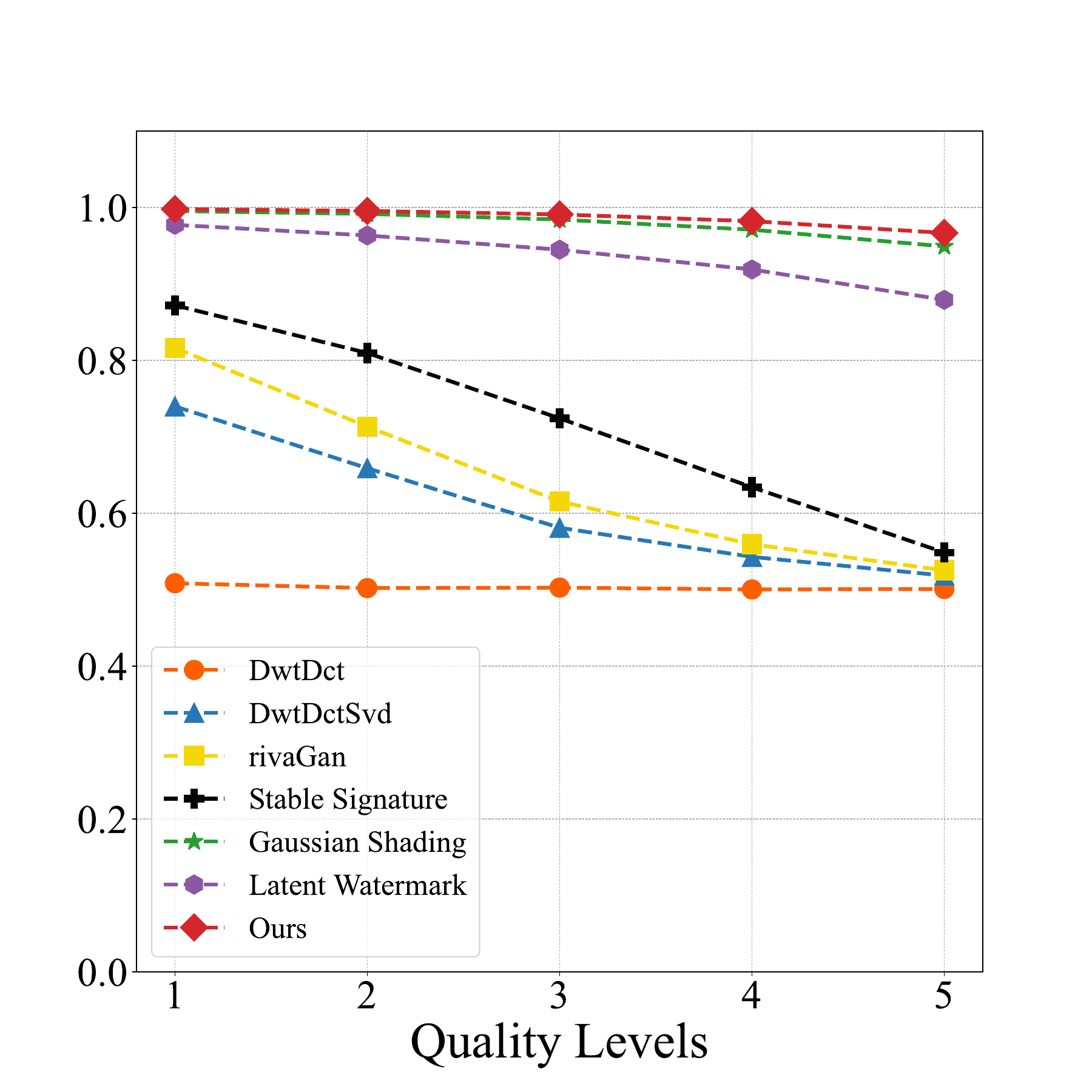}}
    % \hfill
	\subfloat[\label{fig:cheng2020_attn}]{\includegraphics[width = 0.24\textwidth]{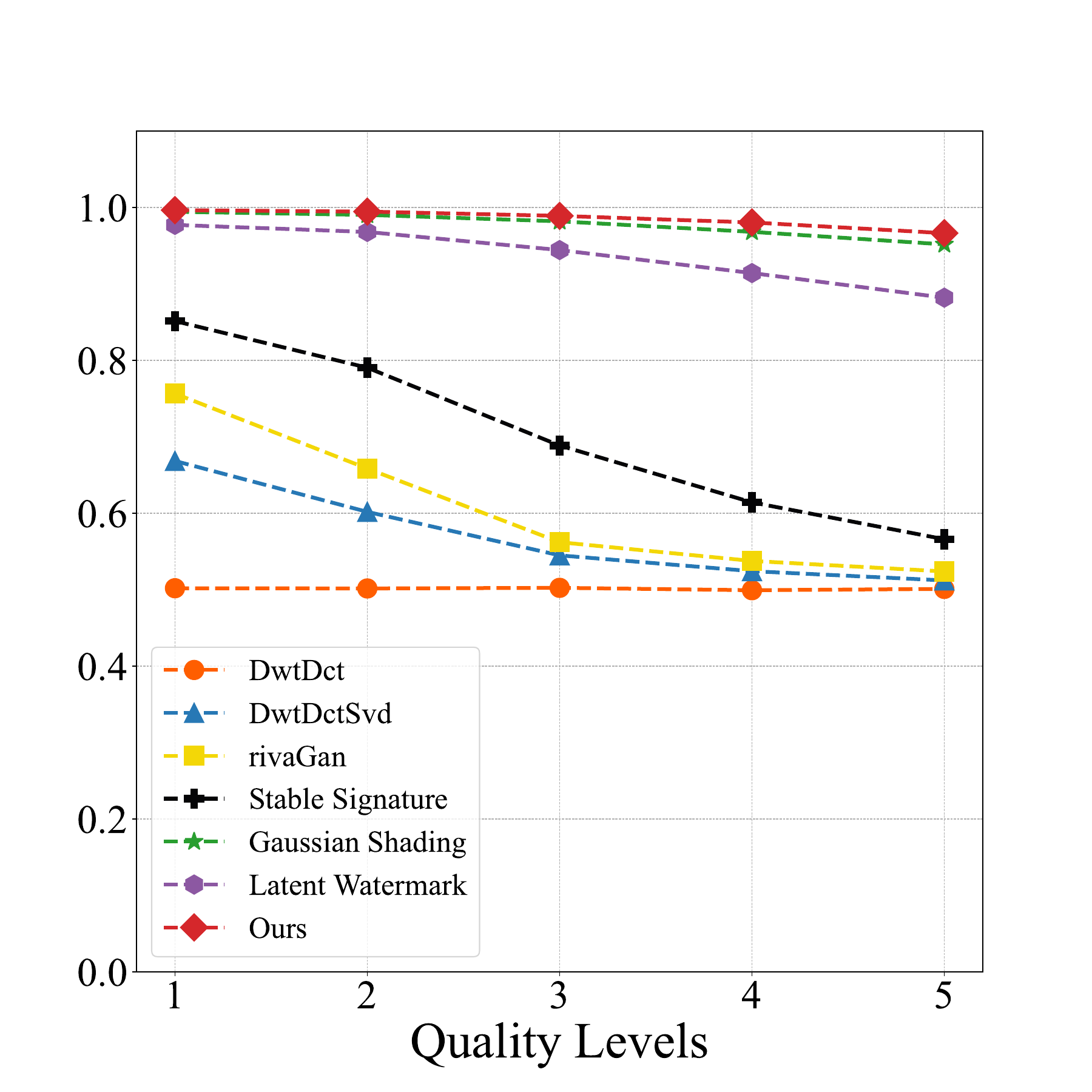}}
\caption{
Performance comparison of different watermarking methods under various attack types and intensities, evaluated using bit accuracy as the performance metric. The plots illustrate the robustness of methods against (a) median filtering, (b) JPEG compression, (c) Gaussian blur, (d) Gaussian noise, (e) salt-and-pepper noise, (f) resizing and restoring, (g) Regen-VAE-A, and (h) Regen-VAE-B.}
\label{fig:attack_result}
\end{figure*}

\section{Experiments}
\subsection{Experimental Settings}
\subsubsection{Implementation details}
We evaluated the efficacy of TraceMark-LDM by employing Stable Diffusion v2.1 as the foundational model. The resolution of the generated watermarked images is configured to \(512 \times 512\), with the latent space dimensions defined as \(4 \times 64 \times 64\). Prompts from the Stable-Diffusion-Prompt repository are used in conjunction with the DPMsolver \cite{Lu2022DPMSolverAF} multistep scheduler, utilizing a 50-step sampling process and a default guidance scale of 7.5. During the watermark extraction phase, DDIM inversion is performed with the same number of inversion steps, null prompts, and a guidance factor of 1. 

To fine-tune the encoder, we simulate various distortion environments by applying different attacks, such as median filter (A), JPEG compression (B), Gaussian blur (C), Gaussian noise (D), salt and pepper noise (E), and resize and restore (F), to 200 clean images. The parameter settings for these attacks, as shown in Table \ref{tab:attack method}, represent a range of common distortions. The training process uses the AdamW optimizer with a learning rate of \(1 \times 10^{-6}\), and the loss function weight coefficient 
\(\lambda\) is set to 1. The model is trained for 100 epochs.

All experiments are performed with PyTorch 2.1.1. The GPU is a NVIDIA RTX 3090Ti with a memory size of 24GB.

\begin{table}[ht]
    \caption{Attack Method and Parameter Ranges}
    \label{tab:attack method}
    \centering
    \begin{tabular}{ccc}
        ID & Attack Method & Parameter Ranges \\ \hline
        A & Median Filter & Kernel size: 3 - 19 \\ 
        B & JPEG Compression & Quality factor: 10 - 90 \\ 
        C & Gaussian Blur & Radius: 2.0 - 10.0 \\ 
        D & Gaussian Noise & Standard deviation: 0.05 - 0.25 \\ 
        E & Salt and Pepper Noise & Probability: 0.05 - 0.4 \\ 
        F & Resize and Restore & Scaling factor: 0.1 - 0.9 \\ 
        G & Regen-VAE-A & Quality level: 1 - 5 \\ 
        H & Regen-VAE-B & Quality level: 1 - 5 \\ 
        I & Regen-Diffusion &  Noise steps: 300 - 700 \\ 
    \end{tabular}
\end{table}

\begin{table*}[ht]
\centering
\caption{Bit accuracy across different methods under varying attack levels.}
\label{tab:detailed_comparison}
\begin{tabular}{cccccccccc}
\toprule
\multicolumn{2}{c}{\multirow{2}{*}{\textbf{Attack}}} & \multicolumn{7}{c}{\textbf{Methods}} \\
\cmidrule(lr){3-10}
 && \cite{cox2007digital} & \cite{cox2007digital} & \cite{Zhang2019RobustIV} &  \cite{fernandez2023stable} &\cite{NEURIPS2023_b54d1757} &  \cite{yang2024gaussian} &  \cite{meng2024latent} & \textbf{Ours} \\

\midrule
\multirow{5}{*}{\textbf{MedFilter}} 
    & \textbf{3}  & 0.371/0.6265  & 0.999/0.9972  & 0.950/0.9882  & 0.961/0.9505  & 1.000/-  & 1.000/0.9999  & 1.000/0.9970   & \textbf{1.000/1.0000} \\
    & \textbf{7}  & 0.001/0.5203  & 0.996/0.9302  & 0.868/0.9693  & 0.000/0.6473  & 1.000/-  & 1.000/0.9988  & 1.000/0.9792  & \textbf{1.000/0.9992} \\
    & \textbf{11} & 0.000/0.5093  & 0.966/0.7965  & 0.518/0.8881  & 0.000/0.5222  & 0.997/-  & 1.000/0.9837  & 1.000/0.9209  & \textbf{1.000/0.9944} \\
    & \textbf{15} & 0.000/0.5063  & 0.691/0.6861  & 0.072/0.7391  & 0.000/0.4739  & 1.000/-  & 0.997/0.9151  & 1.000/0.8234  & \textbf{1.000/0.9829} \\
    & \textbf{19} & 0.000/0.5038  & 0.260/0.6133  & 0.007/0.6448  & 0.000/0.4500  & 1.000/-  & 0.972/0.8249  & 1.000/0.7190  & \textbf{0.999/0.9607} \\
\midrule

\multirow{5}{*}{\textbf{JPEG}} 
    & \textbf{90} & 0.042/0.5563 & 0.996/0.9821 & 0.933/0.9848 & 0.969/0.9446 & 1.000/- & 1.000/0.9999 & 1.000/0.9964 & \textbf{1.000/0.9999} \\
    & \textbf{70} & 0.000/0.5159 & 0.994/0.8896 & 0.839/0.9636 & 0.849/0.8932 & 1.000/- & 1.000/0.9985 & 1.000/0.9928 & \textbf{1.000/0.9993} \\
    & \textbf{50} & 0.001/0.5073 & 0.861/0.7265 & 0.630/0.9193 & 0.656/0.8561 & 1.000/- & 1.000/0.9965 & 1.000/0.9868 & \textbf{1.000/0.9985} \\
    & \textbf{30} & 0.000/0.5017 & 0.008/0.5342 & 0.222/0.8279 & 0.302/0.7938 & 1.000/- & 1.000/0.9902 & 1.000/0.9702 & \textbf{1.000/0.9962} \\
    & \textbf{10} & 0.000/0.5019 & 0.000/0.4912 & 0.000/0.5866 & 0.000/0.5727 & 0.997/- & 1.000/0.9447 & 1.000/0.8780 & \textbf{1.000/0.9766} \\
\midrule

\multirow{5}{*}{\textbf{GauBlur}} 
    & \textbf{2}  & 0.005/0.5245  & 0.988/0.8993  & 0.872/0.9723  & 0.000/0.5456  & 1.000/-  & 1.000/0.9996  & 1.000/0.9882 & \textbf{1.000/0.9999} \\
    & \textbf{4}  & 0.001/0.5034  & 0.530/0.6631  & 0.194/0.7943  & 0.000/0.4165  & 0.999/-  & 1.000/0.9843  & 1.000/0.9221 & \textbf{1.000/0.9967} \\
    & \textbf{6}  & 0.000/0.4999  & 0.102/0.5662  & 0.000/0.5984  & 0.000/0.4083  & 0.998/-  & 1.000/0.9194  & 0.997/0.8160 & \textbf{1.000/0.9660} \\
    & \textbf{8}  & 0.000/0.4995  & 0.007/0.5333  & 0.000/0.5359  & 0.000/0.3967  & 0.998/-  & 0.990/0.8263  & 0.756/0.6877 & \textbf{0.999/0.9152} \\
    & \textbf{10} & 0.000/0.4995  & 0.002/0.5164  & 0.000/0.5175  & 0.000/0.3927  & 0.999/-  & 0.954/0.7486  & 0.060/0.5923 & \textbf{0.996/0.8491} \\
\midrule

\multirow{5}{*}{\textbf{GauNoise}} 
    & \textbf{0.05} & 0.374/0.6329 & 0.893/0.8227 & 0.465/0.8589 & 0.324/0.7409 & 0.890/- & 0.995/0.9546 & 0.938/0.8245 & \textbf{1.000/0.9968} \\
    & \textbf{0.10} & 0.000/0.4699 & 0.000/0.5243 & 0.046/0.7142 & 0.000/0.5386 & 0.772/- & 0.949/0.8617 & 0.614/0.6869 & \textbf{1.000/0.9882} \\
    & \textbf{0.15} & 0.000/0.4965 & 0.000/0.5011 & 0.000/0.6386 & 0.000/0.5236 & 0.672/- & 0.826/0.7647 & 0.266/0.6155 & \textbf{1.000/0.9738} \\
    & \textbf{0.20} & 0.000/0.5012 & 0.000/0.5022 & 0.000/0.6066 & 0.000/0.5171 & 0.487/- & 0.651/0.6955 & 0.059/0.5702 & \textbf{0.998/0.9466} \\
    & \textbf{0.25} & 0.000/0.5022 & 0.000/0.4990 & 0.000/0.5950 & 0.000/0.5308 & 0.338/- & 0.407/0.6400 & 0.005/0.5396 & \textbf{0.993/0.8931} \\
\midrule

\multirow{5}{*}{\textbf{S\&P Noise}} 
    & \textbf{0.05} & 0.065/0.5724 & 0.000/0.5193 & 0.260/0.8553 & 0.021/0.6742 & 0.994/- & 0.998/0.9359 & 0.788/0.7319 & \textbf{1.000/0.9997} \\
    & \textbf{0.10} & 0.000/0.5073 & 0.000/0.4984 & 0.029/0.7576 & 0.000/0.5530 & 0.992/- & 0.995/0.9141 & 0.733/0.7081 & \textbf{1.000/0.9994} \\
    & \textbf{0.20} & 0.000/0.4954 & 0.000/0.4999 & 0.001/0.6705 & 0.000/0.5209 & 0.978/- & 0.994/0.8711 & 0.536/0.6685 & \textbf{1.000/0.9987} \\
    & \textbf{0.30} & 0.000/0.4969 & 0.000/0.5000 & 0.000/0.6306 & 0.000/0.5441 & 0.981/- & 0.971/0.8305 & 0.294/0.6285 & \textbf{1.000/0.9970} \\
    & \textbf{0.40} & 0.000/0.4987 & 0.000/0.5010 & 0.000/0.6082 & 0.000/0.5427 & 0.929/- & 0.932/0.7909 & 0.111/0.5973 & \textbf{1.000/0.9942} \\
\midrule

\multirow{5}{*}{\textbf{Resize}} 
    & \textbf{0.9} & 0.791/0.7662 & 0.997/0.9959 & 0.953/0.9893 & 0.982/0.9745 & 1.000/- & 1.000/1.0000 & 1.000/0.9975 & \textbf{1.000/1.0000} \\
    & \textbf{0.7} & 0.564/0.6731 & 0.997/0.9940 & 0.951/0.9888 & 0.918/0.9354 & 1.000/- & 1.000/0.9999 & 1.000/0.9974 & \textbf{1.000/1.0000} \\
    & \textbf{0.5} & 0.246/0.5976 & 0.997/0.9809 & 0.939/0.9860 & 0.261/0.7888 & 1.000/- & 1.000/0.9996 & 1.000/0.9956 & \textbf{1.000/0.9999} \\
    & \textbf{0.3} & 0.004/0.5224 & 0.992/0.8908 & 0.878/0.9738 & 0.001/0.5992 & 1.000/- & 1.000/0.9988 & 1.000/0.9853 & \textbf{1.000/0.9996} \\
    & \textbf{0.1} & 0.000/0.5007 & 0.071/0.5655 & 0.000/0.5942 & 0.000/0.4019 & 0.998/- & 0.996/0.8339 & 0.983/0.7641 & \textbf{0.999/0.9052} \\
\midrule

\multirow{5}{*}{\textbf{Regen-VAE-A}} 
    & \textbf{5} & 0.001/0.5084 & 0.866/0.7398 & 0.267/0.8163 & 0.708/0.8720 & 0.996/- & 1.000/0.9954 & 1.000/0.9774 & \textbf{1.000/0.9978} \\
    & \textbf{4} & 0.000/0.5021 & 0.558/0.6589 & 0.035/0.7133 & 0.466/0.8096 & 0.996/- & 1.000/0.9916 & 1.000/0.9634 & \textbf{1.000/0.9955} \\
    & \textbf{3} & 0.000/0.5027 & 0.043/0.5810 & 0.001/0.6153 & 0.155/0.7242 & 0.987/- & 0.999/0.9842 & 1.000/0.9448 & \textbf{1.000/0.9909} \\
    & \textbf{2} & 0.000/0.5003 & 0.000/0.5429 & 0.000/0.5594 & 0.008/0.6340 & 0.994/- & 0.999/0.9709 & 0.997/0.9187 & \textbf{1.000/0.9821} \\
    & \textbf{1} & 0.000/0.5009 & 0.000/0.5184 & 0.000/0.5253 & 0.000/0.5486 & 0.990/- & 0.997/0.9491 & 0.995/0.8792 & \textbf{1.000/0.9667} \\
\midrule

\multirow{5}{*}{\textbf{Regen-VAE-B}} 
    & \textbf{5} & 0.000/0.5019 & 0.600/0.6684 & 0.083/0.7565 & 0.629/0.8514 & 0.999/- & 1.000/0.9944 & 1.000/0.9772 & \textbf{1.000/0.9965} \\
    & \textbf{4} & 0.000/0.5016 & 0.166/0.6018 & 0.004/0.6585 & 0.399/0.7904 & 0.999/- & 0.999/0.9903 & 1.000/0.9680 & \textbf{1.000/0.9946} \\
    & \textbf{3} & 0.000/0.5024 & 0.001/0.5449 & 0.000/0.5621 & 0.068/0.0686 & 0.998/- & 0.999/0.9819 & 0.998/0.9443 & \textbf{1.000/0.9890} \\
    & \textbf{2} & 0.000/0.4993 & 0.000/0.5242 & 0.000/0.5376 & 0.010/0.6142 & 0.995/- & 0.998/0.9681 & 0.997/0.9142 & \textbf{1.000/0.9804} \\
    & \textbf{1} & 0.000/0.5010 & 0.000/0.5121 & 0.000/0.5239 & 0.000/0.5660 & 0.989/- & 0.996/0.9517 & 0.997/0.8820 & \textbf{0.999/0.9664} \\
\midrule

\multirow{5}{*}{\textbf{Regen-Diffusion}} 
    & \textbf{300} & 0.000/0.5010 & 0.005/0.5072 & 0.000/0.5478 & 0.000/0.4660 & 0.978/- & 1.000/0.9527 & 1.000/0.8123 & \textbf{1.000/0.9702} \\
    & \textbf{400} & 0.000/0.5003 & 0.000/0.5036 & 0.000/0.5445 & 0.000/0.4585 & 0.939/- & 1.000/0.9125 & 0.992/0.7478 & \textbf{1.000/0.9376} \\
    & \textbf{500} & 0.000/0.5003 & 0.000/0.5000 & 0.000/0.5409 & 0.000/0.4599 & 0.898/- & 0.999/0.8545 & 0.870/0.6916 & \textbf{0.999/0.8860} \\
    & \textbf{600} & 0.000/0.5003 & 0.000/0.4992 & 0.000/0.5418 & 0.000/0.4608 & 0.773/- & 0.989/0.7861 & 0.356/0.6422 & \textbf{0.995/0.8193} \\
    & \textbf{700} & 0.000/0.5004 & 0.000/0.4980 & 0.000/0.5384 & 0.000/0.4556 & 0.570/- & 0.923/0.7155 & 0.040/0.5991 & \textbf{0.954/0.7416} \\
    
\bottomrule
\end{tabular}

\end{table*}

\subsubsection{Baseline Methods}
Several post-generation watermarking techniques, including DwtDct \cite{cox2007digital}, DwtDctSvd \cite{cox2007digital}, and RivaGAN \cite{Zhang2019RobustIV}, are compared with our TraceMark-LDM. For in-generationwatermarking methods, we employ Stable Signature \cite{fernandez2023stable} and Latent Watermark \cite{meng2024latent}. For initial noise sampling watermarking methods, we adopt Tree-Ring \cite{NEURIPS2023_b54d1757} and Gaussian Shading \cite{yang2024gaussian}. To ensure a fair comparison, we standardize the watermark capacity of TraceMark-LDM to 256 bits.

\subsection{Evaluation Metrics}
\subsubsection{Watermark Accuracy}
In order to evaluate the performance of watermark detection, we adopt the True Positive Rate at a False Positive Rate of \(10^{-6}\) (TPR@\(10^{-6}\) FPR) as the primary metric. TPR@\(10^{-6}\) FPR is defined as the detection accuracy calculated using the corresponding threshold that ensures the theoretical false positive rate remains below \(10^{-6}\). We assume that the matching or mismatching between the injected watermark bits and the extracted watermark bits are independent and identically distributed variables following a Bernoulli distribution with a probability parameter of 0.5. For comparison methods utilizing watermark lengths of 32, 48, and 256 bits, the respective thresholds are set to 30, 41, and 167 bits. The detection accuracy is computed using the following formula:
\begin{equation}
\text{TPR} = \frac{1}{N} \sum_{n=1}^{N} \mathbb{I}\left(\sum_{k=1}^{K} \left( m_{k,n} \equiv m_{k,n}' \right) > \tau \right)
\label{eq:TPR}
\end{equation}

where \(N\) is the total number of samples, \(K\) is the watermark length, \(m_{k,n}\) and \(m_{k,n}'\) are the \(k\)-th watermark bits injected and extracted for the \(n\)-th sample respectively, \(\tau\) is the threshold, and \(\mathbb{I}\{\cdot\}\) is the indicator function, which equals 1 if the condition within the braces is satisfied, and 0 otherwise.

In the context of attribution watermarking, we use the average bit accuracy of all extracted watermark samples as the performance metric to evaluate our method. The specific formula is defined as follows:
\begin{equation}
\text{Bit Accuracy} = \frac{1}{N \cdot K} \sum_{n=1}^{N} \sum_{k=1}^{K} \left( m_{k,n} \equiv m_{k,n}' \right)
\label{eq:ACC}
\end{equation}

\subsubsection{Image Quality}
To evaluate the quality of watermarked images, we utilize two widely recognized metrics: FID and CLIP-Score.

To analyze the potential impact of watermark embedding on the model's performance, we conduct a statistical evaluation using a \textit{t}-test. Specifically, we define the null hypothesis (\(H_0\)) and the alternative hypothesis (\(H_1\)) as \( H_0 : \mu_s = \mu_0 \) and \( H_1 : \mu_s \neq \mu_0 \), where \(\mu_s\) and \(\mu_0\) represent the average FID or CLIP-Score for multiple sets of watermarked and watermark-free images, respectively. The \textit{t}-test is used to determine whether the watermark embedding introduces statistically significant performance differences. A smaller \textit{t}-value suggests a higher likelihood of accepting \(H_0\), indicating no significant performance impact. Conversely, if the \textit{t}-value exceeds a predefined threshold, \(H_0\) is rejected, suggesting that watermark embedding may influence the model's performance.

\subsection{Comparison with Baselines}
\subsubsection{Robustness Evaluation against Image Processing}
In our experiments, we comprehensively evaluated the robustness of various watermarking methods in extracting watermarks from 1000 generated watermarked images under different image distortions. The specific parameter settings for each distortion method are provided in Table \ref{tab:attack method}.  

As shown in Table \ref{tab:detailed_comparison}, the data before and after the slash represent the TPR and bit accuracy, respectively. The results demonstrate that our method consistently achieved the highest watermark extraction accuracy across all distortion scenarios. Notably, in the Gaussian noise scenario, our method attained an average detection accuracy of 99.8\%, surpassing the best-performing baseline, Gaussian Shading \cite{yang2024gaussian}, by 23.2\%. Fig. \ref{fig:attack_result} illustrates the performance trends of each method as the attack intensity increases. Even under salt-and-pepper noise, our method consistently maintained a high bit accuracy, while the performance of other methods significantly degraded. This result is attributed to our fine-tuned LDM encoder, which adapts well to noise addition and denoising attacks, ensuring that the decoded latent variables closely match the original latent variables, thereby significantly improving watermark extraction accuracy. These results validate the robustness and wide applicability of TraceMark-LDM under diverse and challenging conditions.
\begin{figure}[ht]
\centering
\includegraphics[width=\linewidth]{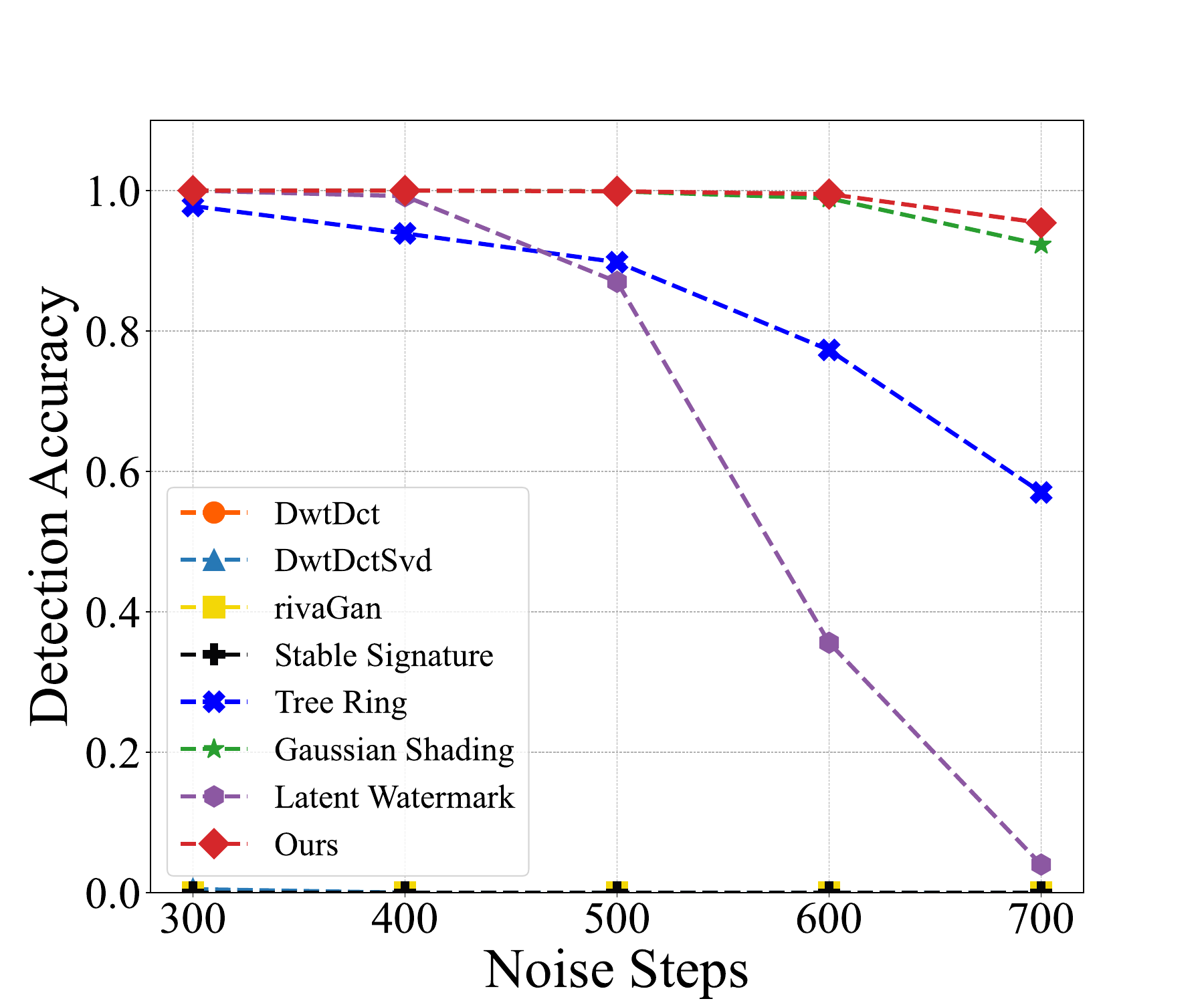}
\caption{Detection accuracy of watermarking methods under diffusion-based regeneration attacks.}
\label{fig:diffusion_based_attack}
\end{figure}
\subsubsection{Robustness Evaluation against VAE-Based Regeneration Attacks}

In our experiments, to evaluate the robustness of watermarking methods against VAE-based regeneration attacks, we used two VAE-based image compression models from the compressAI library: Bmshj2018 \cite{Ball2018VariationalIC} and Cheng2020 \cite{Cheng2020LearnedIC}, referred to as Regen-VAE-A and Regen-VAE-B, respectively. The compression factors were set to [1, 2, 3, 4, 5], where a smaller compression factor indicates more severe image compression, resulting in a decrease in watermark extraction accuracy. As shown in Table \ref{tab:detailed_comparison}, although most methods can still extract watermarks effectively when the compression factor is 5, watermark extraction accuracy significantly declines as the compression intensity increases, and many methods fail to detect the watermark at higher compression factors. While Gaussian Shading \cite{yang2024gaussian}, TreeRing\cite{NEURIPS2023_b54d1757}, and Latent Watermark \cite{meng2024latent} methods can resist the compression effects of these two VAE models to some extent, their performance is still inferior to the proposed method.

\subsubsection{Robustness Evaluation against Diffusion-Based Regeneration Attacks}
Diffusion-based regeneration attacks \cite{zhao2023invisible} introduce noise and apply denoising processes to watermarked images using diffusion models, aiming to disrupt watermark information. Research indicates that most existing generative model watermarks exhibit poor robustness against diffusion-based attacks. To evaluate the performance of our method under such attacks, we selected five different diffusion step sizes: 300, 400, 500, 600, and 700. As the diffusion step size increases, the deviation between the image and the original image becomes larger, typically leading to performance degradation. As shown in Fig. \ref{fig:diffusion_based_attack}, our method demonstrates the strongest resistance to this type of attack, which can be attributed to our grouping and rearrangement strategy. This strategy independently handles elements with small absolute values in the latent variables, effectively mitigating the impact of image deviations caused by the diffusion attack. In contrast, other post-generation watermarking methods fail to resist such attacks, causing detection accuracy to approach 50\%. For the StableSignature method \cite{fernandez2023stable}, the watermark is completely erased during the attack process due to the use of a non-watermark decoder.

\subsubsection{Image Quality Evaluation}
\begin{figure*}[htbp]
    \centering
    \includegraphics[width=\textwidth]{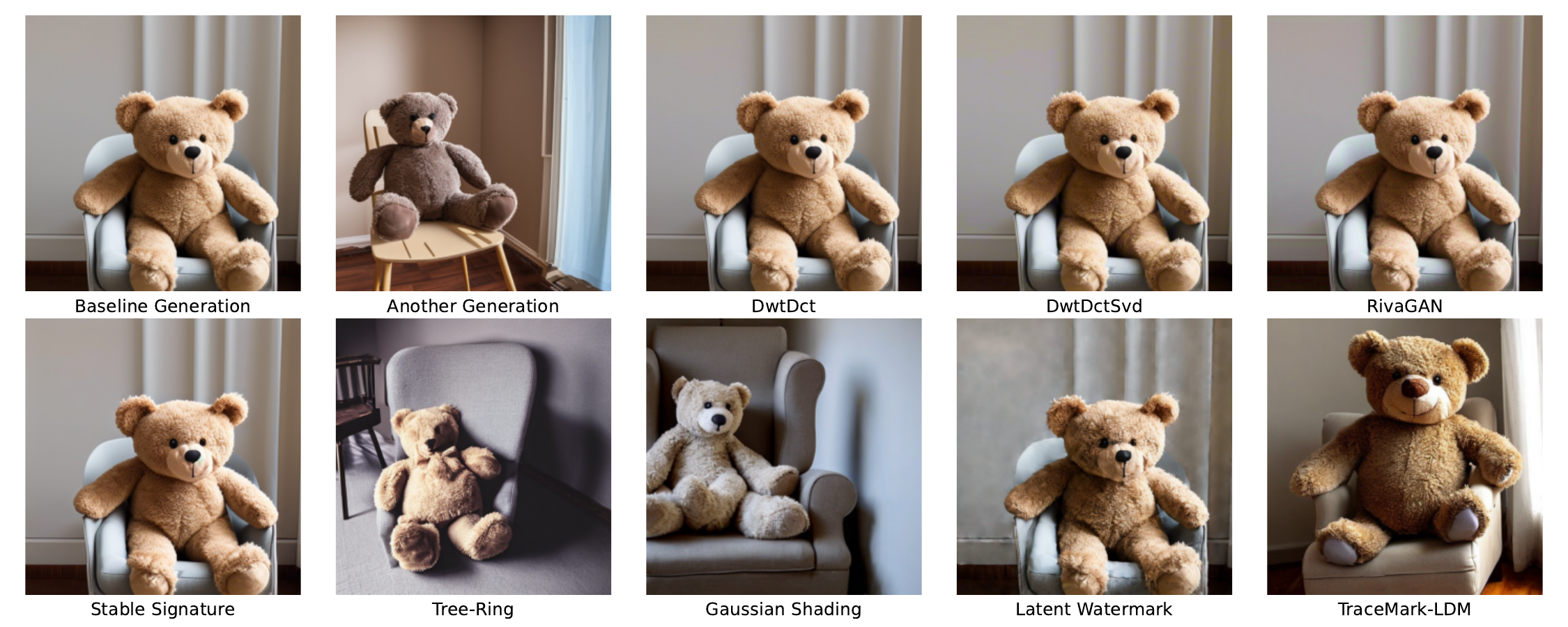}
    \caption{Examples of different watermarking methods applied to images generated with the prompt: ``A teddy bear sitting on a chair inside a room." \textbf{Baseline Generation} refers to the standard generated image without any watermark, serving as the reference baseline. \textbf{Another Generation} refers to another independent image generated with the same prompt, showcasing the inherent diversity of the model. } 
    \label{fig:visual_legend}
\end{figure*}

To quantitatively evaluate the quality of watermarked images, we calculated the FID and CLIP scores of various watermarking methods relative to the baseline model. For FID evaluation, 10 sets of 5,000 images were generated from the COCO dataset. For CLIP score evaluation, 10 sets of 1,000 images were generated using prompts from the Stable-Diffusion-Prompt. A \textit{t}-test was then conducted to assess the significance of differences in FID and CLIP scores between watermarked and non-watermarked images. Fig. \ref{fig:visual_legend} shows visualization examples of different methods. Notably, the image with the Latent Watermark \cite{meng2024latent} exhibits significant distortion.

\begin{table}[ht]
\caption{Image quality evaluation across different methods, highlighting the top two results with the lowest t-values.}
    \label{tab:image_quality}
    \centering
    \begin{tabular}{lcc}
    \toprule
        Method & FID (\textit{t}-value $\downarrow$) & CLIP-Score (\textit{t}-value $\downarrow$) \\ 
        \midrule
        Stable Diffusion & 24.90$_{.02}$ & 0.3647$_{.0011}$ \\ 
        DwtDct \cite{cox2007digital} & 24.76$_{.02}$ (2.189) & 0.3629$_{.0011}$ (3.608) \\ 
        DwtDctSvd \cite{cox2007digital} & 24.27$_{.02}$ (9.559) & 0.3622$_{.0010}$ (5.059) \\ 
        rivaGan \cite{Zhang2019RobustIV} & 24.02$_{.02}$ (13.81) & 0.3624$_{.0011}$ (4.629) \\ 
        Stable Signature \cite{fernandez2023stable} & 25.49$_{.10}$ (10.03) & 0.3635$_{.0011}$ (2.347) \\ 
        Tree Ring \cite{NEURIPS2023_b54d1757} & 25.34$_{.02}$ (6.398) & 0.3631$_{.0007}$ (3.632) \\ 
        Gaussian Shading \cite{yang2024gaussian} & 24.88$_{.07}$ (\textbf{0.169}) & 0.3645$_{.0006}$ (\textbf{0.563}) \\ 
        Latent watermark \cite{meng2024latent} & 27.12$_{.16}$ (32.51) & 0.3583$_{.0009}$ (13.99) \\
        Ours & 24.96$_{.03}$ (\textbf{0.773}) & 0.3649$_{.0010}$ (\textbf{0.372}) \\ 
    \bottomrule
    \end{tabular}
    
\end{table}
As demonstrated by the results in brackets in Table \ref{tab:image_quality}, the \textit{t}-values for TraceMark-LDM are all below the threshold of 2.101 specified in \cite{yang2024gaussian}. This indicates that TraceMark-LDM has negligible impact on the quality and semantic consistency of synthesized images during watermark embedding. In contrast, the \textit{t}-values of other methods, except for Gaussian Shading, exceed the threshold. This is because these methods modify certain process variables during image generation, which negatively affects image quality. For the CLIP score, our method achieved the best performance, showing the smallest deviation from the baseline. Additionally, we observed that some post-generation methods consistently resulted in lower FID scores. This phenomenon may be attributed to the watermarking process smoothing the texture of images, which can lead to visually more coherent results.

\begin{table}[htbp]
\caption{Ablation study of different components of our proposed method under various adversarial attacks.}
\label{tab:ablation_comparison}
\centering
\setlength{\tabcolsep}{1mm}
\begin{tabular}{cccccc}
\toprule
\multicolumn{2}{c}{\multirow{2}{*}{\textbf{Attack}}} & \multicolumn{4}{c}{\textbf{Methods}} \\
\cmidrule(lr){3-6}
 &  & \textbf{Ours} & \textbf{w/o FT, with GR} & \textbf{w/o FT, w/o S} & w/o FT, with R \\
\midrule

\multirow{5}{*}{\textbf{A}} 
    & \textbf{3}   & 1.000/1.0000 & 1.000/1.0000 & 1.000/1.0000 & 1.000/1.0000\\
    & \textbf{7}   & 1.000/0.9992 & 1.000/0.9997 & 1.000/0.9996 & 1.000/0.9989\\
    & \textbf{11}  & 1.000/0.9944 & 1.000/0.9929 & 1.000/0.9912 & 1.000/0.9849\\
    & \textbf{15}  & 1.000/0.9829 & 0.999/0.9410 & 1.000/0.9345 & 0.996/0.9161\\
    & \textbf{19}  & 0.999/0.9607 & 0.987/0.8558 & 0.982/0.8479 & 0.969/0.8240\\
\midrule

\multirow{5}{*}{\textbf{B}} 
    & \textbf{90}  & 1.000/0.9999 & 1.000/1.0000 & 1.000/1.0000 & 1.000/0.9999\\
    & \textbf{70}  & 1.000/0.9993 & 1.000/0.9993 & 1.000/0.9992 & 1.000/0.9984\\
    & \textbf{50}  & 1.000/0.9985 & 1.000/0.9982 & 1.000/0.9980 & 1.000/0.9990\\
    & \textbf{30}  & 1.000/0.9962 & 1.000/0.9948 & 1.000/0.9940 & 0.999/0.9901\\
    & \textbf{10}  & 1.000/0.9766 & 0.997/0.9608 & 1.000/0.9587 & 0.998/0.9420\\
\midrule

\multirow{5}{*}{\textbf{C}} 
    & \textbf{2}   & 1.000/0.9999 & 1.000/0.9999 & 1.000/0.9999 & 1.000/0.9997\\
    & \textbf{4}   & 1.000/0.9967 & 1.000/0.9953 & 1.000/0.9939 & 1.000/0.9889\\
    & \textbf{6}   & 1.000/0.9660 & 1.000/0.9509 & 1.000/0.9458 & 1.000/0.9260\\
    & \textbf{8}   & 0.999/0.9152 & 0.998/0.8618 & 0.996/0.8547 & 0.992/0.8263\\
    & \textbf{10}  & 0.996/0.8491 & 0.970/0.7810 & 0.967/0.7739 & 0.945/0.7475\\
\midrule

\multirow{5}{*}{\textbf{D}} 
    & \textbf{0.05} & 1.000/0.9968 & 0.997/0.9662 & 0.998/0.9618 & 0.994/0.9504\\
    & \textbf{0.1}  & 1.000/0.9882 & 0.973/0.8796 & 0.960/0.8689 & 0.943/0.8492\\
    & \textbf{0.15} & 1.000/0.9738 & 0.874/0.7907 & 0.860/0.7808 & 0.811/0.7596\\
    & \textbf{0.2}  & 0.998/0.9466 & 0.715/0.7170 & 0.668/0.7077 & 0.640/0.6885\\
    & \textbf{0.25} & 0.993/0.8931 & 0.489/0.6578 & 0.441/0.6501 & 0.387/0.6337\\
\midrule

\multirow{5}{*}{\textbf{E}} 
    & \textbf{0.05} & 1.000/0.9997 & 1.000/0.9545 & 1.000/0.9481 & 0.996/0.9293\\
    & \textbf{0.1}  & 1.000/0.9994 & 1.000/0.9361 & 0.999/0.9276 & 0.991/0.9048\\
    & \textbf{0.2}  & 1.000/0.9987 & 1.000/0.9008 & 0.993/0.8892 & 0.986/0.8626\\
    & \textbf{0.3}  & 1.000/0.9970 & 0.989/0.8626 & 0.989/0.8512 & 0.965/0.8219\\
    & \textbf{0.4}  & 1.000/0.9942 & 0.979/0.8243 & 0.963/0.8107 & 0.936/0.7828\\
\midrule

\multirow{5}{*}{\textbf{F}} 
    & \textbf{0.9}  & 1.000/1.0000 & 1.000/1.0000 & 1.000/1.0000 & 1.000/1.0000\\
    & \textbf{0.7}  & 1.000/1.0000 & 1.000/1.0000 & 1.000/1.0000 & 1.000/1.0000\\
    & \textbf{0.5}  & 1.000/0.9999 & 1.000/1.0000 & 1.000/1.0000 & 1.000/0.9999\\
    & \textbf{0.3}  & 1.000/0.9996 & 1.000/0.9996 & 1.000/0.9996 & 1.000/0.9988\\
    & \textbf{0.1}  & 0.999/0.9052 & 1.000/0.8698 & 1.000/0.8615 & 0.992/0.8364\\
\bottomrule
\end{tabular}
\end{table}

\subsection{Ablation Studies}
We conducted a series of ablation studies to validate the significance of fine-tuning and group rearrangement, and to evaluate the impact of various parameters on our method, including watermark length, inversion step, and sampling method.
\subsubsection{Effectiveness of Fine-Tuning}
To confirm the effectiveness of the fine-tuned encoder, we applied the same attack scenarios as others on images generated by unfine-tuned TraceMark-LDM. As shown in Table \ref{tab:ablation_comparison}, `w/o FT' refers to the performance without fine-tuning. The results indicate that, even without fine-tuning, our method surpasses existing benchmarks. However, the performance of the fine-tuned TraceMark-LDM achieves higher detection and bit accuracies than those of the unfine-tuned version. In the case of Gaussian noise with a standard deviation of 0.25, where image distortion is relatively large, the fine-tuned version shows 50.4\% and 23.53\% improvement in detection rate and bit accuracy, respectively. These results underscores the effectiveness of fine-tuning in enhancing robustness while highlighting the inherent strengths of our method's design. Consequently, our method ensures reliable performance across various attacks, regardless of fine-tuning, affirming its potential in real-world applications.

\subsubsection{Effectiveness of Group Rearrangement}
To validate the effectiveness of our Group Rearrangement strategy designed for elements with smaller absolute values, we performed performance tests after discarding these elements. As shown in Table \ref{tab:ablation_comparison}, `with GR' indicates that the Group Rearrangement (GR) strategy is applied to process elements with small absolute values; `w/o S' indicates that the watermark embedding stage does not use elements with small absolute values; `with R' indicates that the Rearrangement (R) strategy is employed to address elements with small absolute values. The results show that even after these elements are removed, the performance is improved compared to the strategy of using rearrangement for all elements, which means that this part of the elements is easily affected by distortion, leading to an increase in the inversion error, and the element symbols are easily reversed after inversion. Furthermore, by reasonably utilizing these elements through the Group Rearrangement strategy, the watermark extraction accuracy achieved consistent improvements across various distortion scenarios. Although the magnitude of improvement is relatively limited, these results clearly demonstrate the reliability and positive impact of the group rearrangement strategy on watermark extraction.

\subsubsection{Watermark Length}
A larger watermark capacity implies that the method can embed more information. The watermark accuracy was evaluated at watermark lengths $k = 128, 256, \ldots, 4096$, and the experimental results are shown in Table \ref{tab:watermark_length_acc}. From the results, it can be observed that as the watermark length increases, the watermark extraction accuracy gradually decreases. This is because, under the condition of a fixed image resolution, the dimensions of the latent variables are also fixed. A longer watermark length results in fewer repetitions of the watermark during embedding, leading to fewer groups available for majority voting during extraction, which in turn reduces extraction accuracy. However, it is worth noting that when the watermark length is 256 bits, the watermark can still be extracted with 100\% accuracy. Even when the watermark length reaches 4096 bits, the extraction accuracy remains as high as 96.8\% under non-attack conditions. Furthermore, even under various attack scenarios, the average watermark extraction accuracy still reaches 81.9\%.

\begin{table}[htbp]
\centering
\caption{Bit accuracy with different watermark lengths.}
\label{tab:watermark_length_acc}
\begin{tabular}{lcccccc}
\toprule
\multirow{2}{*}{\textbf{Attack}} & \multicolumn{6}{c}{\textbf{Watermark Length}} \\
\cmidrule(lr){2-7}
& \textbf{128} & \textbf{256} & \textbf{512} & \textbf{1024} & \textbf{2048} & \textbf{4096} \\
\midrule
Clean & 1.0000 & 1.0000 & 0.9998 & 0.9983 & 0.9926 & 0.9680 \\
Resize & 1.0000 & 0.9999 & 0.9991 & 0.9956 & 0.9822 & 0.9429 \\
GauNoise & 0.9911 & 0.9738 & 0.9351 & 0.8814 & 0.8069 & 0.7373 \\
GauBlur & 0.9911 & 0.9660 & 0.9086 & 0.8328 & 0.7550 & 0.6879 \\
S\&P Noise & 0.9996 & 0.9987 & 0.9923 & 0.9780 & 0.9401 & 0.8807 \\
MedFilter & 0.9989 & 0.9944 & 0.9747 & 0.9285 & 0.8550 & 0.7758 \\
JPEG & 0.9995 & 0.9985 & 0.9929 & 0.9797 & 0.9448 & 0.8882 \\
Avg & 0.9967 & 0.9886 & 0.9671 & 0.9327 & 0.8807 & 0.8188 \\
\bottomrule
\end{tabular}
\end{table}

\subsubsection{Inversion Step}
Typically, the effectiveness of inversion is maximized when the inversion step size matches the inference step size. However, in practical applications, the exact inference step size used during the generation of watermarked images is often unknown. We tested multiple combinations of inversion and inference step sizes, and the results are shown in Table \ref{tab:inversion steps}. It can be observed that our method is not dependent on the inversion step size. Even when the inversion step size is as low as 10, the extraction accuracy remains nearly 100\%. Moreover, when the inversion step size is set to 50, our method accurately extracts watermark information across all the inference step sizes tested.

\begin{table}[ht]
\caption{Bit accuracy with different inference and inversion steps.}
\label{tab:inversion steps}
\centering
\begin{tabular}{ccccc}
\toprule
\multirow{2}{*}{\textbf{Inference Step}} & \multicolumn{4}{c}{\textbf{Inversion Step}} \\
\cmidrule(lr){2-5}
 & \text{10} & \text{25} & \text{50} & \text{100} \\
\midrule
\text{10}  & 0.9999 & 0.9999 & 1.0000 & 1.0000 \\
\text{25}  & 1.0000 & 1.0000 & 1.0000 & 1.0000 \\
\text{50}  & 1.0000 & 1.0000 & 1.0000 & 1.0000 \\
\text{100} & 1.0000 & 1.0000 & 1.0000 & 1.0000 \\
\bottomrule
\end{tabular}
\end{table}

\subsubsection{Sampling Method}
To evaluate the performance of different sampling methods in watermark extraction accuracy, we tested five sampling methods under both undisturbed and various distortion conditions.  

\begin{table}[htbp]
\caption{Bit accuracy across different sampling methods and attack types.}
\label{tab:detail_sampling_methods}
\centering
\begin{tabular}{lccccc}
\toprule
\multirow{2}{*}{\textbf{Attack}} & \multicolumn{5}{c}{\textbf{Sampling Methods}} \\
\cmidrule(lr){2-6}
& \textbf{DDIM} & \textbf{UniPC} & \textbf{PNDM} & \textbf{DEIS} & \textbf{DPMSolver} \\
\midrule
Clean & 1.0000 & 1.0000 & 1.0000 & 1.0000 & 1.0000 \\
Resize & 0.9999 & 0.9999 & 1.0000 & 0.9999 & 0.9999 \\
GauNoise & 0.9660 & 0.9640 & 0.9720 & 0.9704 & 0.9738 \\
GauBlur & 0.9598 & 0.9549 & 0.9663 & 0.9626 & 0.9660 \\
S\&P Noise & 0.9982 & 0.9977 & 0.9987 & 0.9984 & 0.9987 \\
MedianFilter & 0.9928 & 0.9914 & 0.9945 & 0.9936 & 0.9944 \\
JPEG & 0.9979 & 0.9976 & 0.9985 & 0.9982 & 0.9985 \\
Avg & 0.9858 & 0.9843 & 0.9883 & 0.9872 & 0.9886 \\
\bottomrule
\end{tabular}

\end{table}
As shown in Table \ref{tab:detail_sampling_methods}, all sampling methods demonstrate satisfactory performance in the watermark extraction task, providing users with more choices in practical applications and significantly enhancing the method's practicality and adaptability. Notably, DPMSolver achieves an average watermark extraction accuracy of 98.86\% under different disturbance conditions, showcasing exceptional resistance to interference and verifying its robustness in complex scenarios. These results indicate that our method is compatible with multiple sampling strategies and can maintain highly stable watermark extraction performance across diverse application environments.

\subsubsection{Guidance Scale}
By adjusting the guidance scale, a balance can be achieved between image quality and diversity. To ensure that users can successfully extract watermark information using different guidance scales, we conducted tests with five evenly selected guidance scale values ranging from 2 to 18. As illustrated in Table \ref{tab:guidance_scale}, our method consistently maintains a high level of watermark extraction accuracy across various guidance scale values. This further demonstrates the robustness and broad applicability of our approach.

\begin{table}[htbp]
\caption{Bit accuracy across different guidance scales and attack types.}
\label{tab:guidance_scale}
\centering
\begin{tabular}{lccccc}
\toprule
\multirow{2}{*}{\textbf{Attack}} & \multicolumn{5}{c}{\textbf{Guidance Scale}} \\
\cmidrule(lr){2-6}
& \textbf{2.0} & \textbf{6.0} & \textbf{10.0} & \textbf{14.0} & \textbf{18.0} \\
\midrule
Clean & 1.0000 & 1.0000 & 1.0000 & 0.9998 & 0.9991 \\
Resize & 1.0000 & 1.0000 & 0.9998 & 0.9993 & 0.9982 \\
GauNoise & 0.9969 & 0.9810 & 0.9610 & 0.9406 & 0.9232 \\
GauBlur & 0.9895 & 0.9723 & 0.9539 & 0.9336 & 0.9124 \\
S\&P Noise  & 0.9999 & 0.9991 & 0.9972 & 0.9936 & 0.9894 \\
MedianFilter& 0.9993 & 0.9959 & 0.9909 & 0.9835 & 0.9735 \\
JPEG& 0.9999 & 0.9992 & 0.9970 & 0.9943 & 0.9899 \\
Avg& 0.9971 & 0.9895 & 0.9800 & 0.9691 & 0.9577 \\
\bottomrule
\end{tabular}
\end{table}

\section{Conclusion and Future Work}
In this study, we propose TraceMark-LDM, a general and efficient watermarking approach specially designed for LDMs. Unlike traditional watermarking schemes, our method does not directly embed the watermark into images; Instead, it uses the watermark to guide the rearrangement of the initial noise in the LDMs. 
Experimental results demonstrate that TraceMark-LDM effectively enables the tracing and detection of generated content, enhancing the supervision of synthesized images and the trustworthiness of digital data. Furthermore, The FID and CLIP-Score results achieved by TraceMark-LDM are highly comparable to those of other SOTA models.

In the future, the challenge of provable video generation is an intriguing one that could be usefully explored in our further research. We plan to adapt and refine our methods for application in the video generation models, such as the Diffusion Transformer (DiT) behind SORA.

\section*{Acknowledgment}
This work was supported by the National Natural Science Foundation of China under Grant Number 62471167. The authors acknowledge the Supercomputing Center of Hangzhou Dianzi University for providing computing resources. 

The authors gratefully acknowledge members from Sino-France Joint Laboratory for Digital Media Forensics of Zhejiang Province for their valuable feedback and discussions.

\bibliography{arxiv}

\bibliographystyle{IEEEtran}
\end{document}